\documentclass{article} 
\usepackage[main,final]{neurips_2026}

\usepackage{amsmath,amsfonts,bm}

\def\eqref#1{equation~\ref{#1}}

\def\1{\bm{1}}

\def\vx{{\bm{x}}}
\def\vy{{\bm{y}}}

\def\mA{{\bm{A}}}

\def\mY{{\bm{Y}}}

\DeclareMathAlphabet{\mathsfit}{\encodingdefault}{\sfdefault}{m}{sl}
\SetMathAlphabet{\mathsfit}{bold}{\encodingdefault}{\sfdefault}{bx}{n}

\newcommand\varlist{,\makebox[1em][c]{.\hfil.\hfil.},}

\usepackage{hyperref}
\usepackage{url}
\usepackage{graphicx}
\usepackage{subcaption} 
\usepackage{booktabs}
\usepackage{float}
\usepackage{xcolor}
\usepackage{booktabs}

\title{How to make the most of your masked \\ 
language model for protein engineering}

\author{%
  Calvin McCarter\\
  BigHat Biosciences\thanks{\scriptsize Work done while at BigHat Biosciences. Now at Aureka Biotechnologies. Correspondence to: \texttt{mccarter.calvin@gmail.com}.}\\
\And
Nick Bhattacharya\\
BigHat Biosciences\\
\And
Sebastian W. Ober \\
BigHat Biosciences\\  
\And
Hunter Elliott\\
BigHat Biosciences\\
\texttt{helliott@bighatbio.com}
}

\begin{document}

\maketitle

\begin{abstract}
A plethora of protein language models have been released in recent years. Yet comparatively little work has addressed how to best sample from them to optimize desired biological properties. We fill this gap by proposing a flexible, effective sampling method for masked language models (MLMs), and by systematically evaluating models and methods both in silico and in vitro on actual antibody therapeutics campaigns. Firstly, we propose sampling with stochastic beam search, exploiting the fact that MLMs are remarkably efficient at evaluating the pseudo-perplexity of the entire 1-edit neighborhood of a sequence. Reframing generation in terms of entire-sequence evaluation improves the quality of proposed sequence variants, and enables flexible guidance with multiple optimization objectives. Secondly, we report results from our extensive in vitro head-to-head evaluation for the antibody engineering setting, comprising multiple models and samplers. This reveals that the choice of sampling method can have a substantial impact, motivating future research into this under-explored area.
\end{abstract}

\section{Introduction}

Antibodies are a class of natural immune proteins that have served as the starting point for the most successful class of biologic drugs \citep{carter2018next}, and are thus a focus of efforts to utilize machine learning to accelerate drug discovery.
Developing an antibody therapeutic typically involves first identifying a candidate via animal immunization \citep{kohler1975continuous}, \textit{in vitro} screening libraries \citep{mccafferty1990phage}, or \textit{in silico} ``de novo'' design methods \citep{bennett2024atomically}.  
Second, because candidates usually require substantial improvement, \textit{in vitro} iterative optimization is used to refine the candidate into a valid therapeutic \citep{lu2020development}.

The space of possible antibody sequence mutations is combinatorially vast, and only a small number of mutations are typically advantageous; meanwhile, 
multi-objective optimization experiments are limited to hundreds of sequences and are costly and slow \citep{buchanan2025think}.
Machine learning-guided iterative design improves this process by prioritizing mutated sequences which are likely to yield improvements along desired criteria \citep{ober2025sequence,kuo2026multi,rao2026boat}.
In each round, a language model receives top-performing candidates (seeds), and optionally receives additional sequence scoring functions (such as supervised models trained on sequences previously measured in the lab). 
The generative model, combined with a sampling algorithm, outputs mutated sequences, some of which are selected for the next round of lab measurements.

Many protein language models \citep{madani2020progen,lin2022language,alamdari2023protein,hayes2024simulating,bhatnagar2025scaling} and antibody-specific language models \citep{shuai2021generative,olsen2024addressing,fournier2024protein,turnbull2024p,burbach2025curriculum} have been released in recent years.
Because protein sequences lack causal directionality, and because iterative design frequently involves conditioning on arbitrary fixed subsequences, masked language models (MLMs) rather than causal language models (CLMs) are the typical choice. 
Depending on the goals of the optimization campaign, the language model chosen may be one trained on a large human antibody dataset
\citep[The Observed Antibody Space, OAS;][]{kovaltsuk2018observed}, on a database of antibodies with known structures \citep[enriched for therapeutic antibodies,][]{dunbar2014sabdab}, or on a more generic protein sequence database \citep{uniprot2015uniprot}.
Regardless of the choice of model, the premise is that model likelihoods will positively correlate with the probability of the sequence yielding a valid, functional antibody.

The goal of a sampling algorithm is to elicit from the MLM a large, diverse set of sequences with relatively high model likelihood and some notion of proximity to the seed sequences.
Unfortunately, systematic evaluation of sampling MLM algorithms is essentially non-existent.
Furthermore, previously-adopted MLM sampling algorithms employ mutation-centric iterative sampling procedures, essentially any-order analogues of greedy sampling methods for causal language model generation \citep{hoogeboom2021autoregressive,sahoo2024simple}.
These methods are computationally costly and (as we will show later) empirically tend to elicit unlikely, dysfunctional sequences.

It is also frequently desirable to bias the language model's distribution based on additional scoring functions.
Sometimes these are differentiable functions, such as neural networks trained on binding affinity measurements. 
In other cases these scores, such as the OASis percentile risk score for immunogenic adverse events \citep{prihoda2022biophi} or isoelectric point (predictive of stability and pharmacokinetics quality), are both non-differentiable and require as input a clean (rather than partially-masked) sequence.
Mutation-centric sampling methods poorly handle such scenarios without additional approximations \citep{raghu2025guided} or costly computations \citep{tang2025peptune}.

In this work, we propose to instead take a sequence-centric approach, where we evaluate full sequences according to the MLM, and search sequence space starting from the seed.
We show that this is surprisingly computationally efficient and effective at producing sequences preferred by the MLM and optional scoring functions.
Next, we evaluate a variety of models and algorithms, first \textit{in silico} and then \textit{in vitro} on actual antibody therapeutics programs.
We find that the choice of sampler has a surprisingly large impact, making it a consequential and under-explored design choice alongside model selection.
We also report other interesting findings, such as the fact that ESM-2 \citep{lin2022language}, though trained on generic protein sequences, is still very effective for antibody optimization.

\subsection{Related Work}

\paragraph{CLM sampling algorithms} 
Sampling algorithms for causal language models (CLMs) currently tend to follow the greedy decoding paradigm, making decisions solely based on the distribution for the next token.
Argmax decoding often produces repetitive text, so methods like softmax temperature annealing \citep{ackley1985learning,kirkpatrick1983optimization} and nucleus (top-p) filtering \citep{holtzman2019curious} are widely used to spread out probability mass among the most likely tokens, while eliminating the least likely tokens.
Methods based on (partial) sequence likelihoods, such as beam search \citep{gimpel2013systematic,meister2020if} and stochastic beam search (SBS) \citep{kool2019stochastic} have been proposed, but are less frequently used. 
These approaches have complexity $O(L^2)$, where $L$ is the sequence length.

\paragraph{MLM sampling algorithms} Sampling from masked language models (MLMs) for protein engineering has a very different design space from that of text CLMs.
The chief differences are the lack of a pre-defined left-to-right order, the notion of a seed sequence, and the desired proposal of a diverse batch for subsequent \textit{in silico} filtering and finally \textit{in vitro} measurement. 
Also, sometimes specific regions are constrained to be unchanged, such as the complementarity-determining region (CDR) if binding is satisfactory, or the framework region (FR) if only binding requires further modification.

The dominant paradigm for iterative refinement generation from MLMs (and discrete diffusion models), both in NLP \citep{lee2018deterministic,ghazvininejad2019mask,hoogeboom2021autoregressive} and protein engineering \citep{hayes2024simulating,hopf2026evedesign} is mutation-centric: iteratively mask and re-sample subsets of positions.
Denoising sampling \citep{hoogeboom2021autoregressive} is the closest analogue to CLM greedy decoding, and was adopted for protein optimization by \citet{hayes2024simulating}.
To edit up to $E \le L$ residues, conditioned on the remaining $L-E$ fixed positions, it first masks all $E$ positions then runs unmasking for $E$ iterations. 
Alternatively, Gibbs sampling \citep{geman1984stochastic,ackley1985learning} has been adopted for protein \citep{johnson2021generating} engineering, and tends to propose sequences which are closer to the seed sequence.
Gibbs-like sampling masks-and-unmasks for $E$ iterations; at each iteration $1$ position is masked, while the remaining $L-1$ positions are unmasked.
For both autoregressive denoising sampling and Gibbs-like sampling, the sampling step can be replaced with an argmax; this was previously employed for antibody humanization \citep{gordon2024generative}.
Mutation-centric samplers also require a strategy to choose the next position to decode; for example, the ESM-3 work \citep{hayes2024simulating} optimized proteins via selection by either lowest entropy or max probability.

Mutation-centric MLM samplers have complexity $O(EL^3)$ per generated sequence. 
To obtain $N$ unique sequences, one must run the sampling method at least $N$ times; independent runs are not guaranteed to produce unique sequences.

\paragraph{MLM guidance with scoring functions} Prior work on incorporation of additional scoring functions is also limited.
The ESM-3 work \citep{hayes2024simulating} proposes using derivative-free guidance \citep{li2024derivative} but requires scoring functions to accept unclean (partially-masked) sequences.
Discrete Guidance \citep{nisonoff2024unlocking} similarly requires the predictor to score partially-masked sequences.
Alternatively, \citet{raghu2025guided} does not force scoring functions to evaluate unclean sequences, but to remain compatible with mutation-centric sampling, assumes that the scoring function is a product of experts over positions in the sequence \citep{schiff2024simple}.
One recent work \citep{tang2025peptune} uses Monte Carlo tree search (MCTS) to avoid these constraints by performing a complex search and sampling process that scores many clean sequences to estimate the value of one partially-masked sequence; this is problematic for scores like OASis percentile which are very expensive.

\paragraph{Benchmarking} Benchmarking on sampling for protein optimization is scarce.
The most extensive sampler benchmarking results for protein optimization have been reported in \citep{darmawan2025sampling}.
However, it analyzed sampling methods for a causal protein language model \citep{notin2022tranception}, and it involved only \textit{in silico} evaluation.
The argmax versions of denoising and Gibbs-like sampling were evaluated \textit{in vitro} in \citep{gordon2024generative}; argmax Gibbs was marginally better. 

Code is available at \url{https://github.com/calvinmccarter/mlm4proteins}.

\section{Methods}

\subsection{Background on protein MLMs}

Denote the set of 20 amino acids by $\mathcal{A} = \{a_1, a_2 \varlist  a_{20}\}$. Let $\vx = [x_1 \varlist  x_{L}] \in
\mathcal{A}^L$ denote a protein sequence, representing a sequence of $L$ amino acids with each $x_i \in \mathcal{A}$. A masked language model (MLM) takes an input sequence $\vx$ and outputs an $L \times 20$ matrix $\mY = [\vy_1, \dots, \vy_L]^\top$. 
Each $\vy_i$ can naively be interpreted as approximate log-probabilities over the set of amino acids, with the ``probability'' for each residue at location $i$ given by 
\begin{gather}
    \hat{p}(x_i=a_r|{\vx}_{j \ne i}) \approx \frac{\exp(\mY_{ir}/\tau)}{\sum^{20}_{k=1}\exp(\mY_{ik}/\tau)}= \texttt{softmax}_\tau(\mY_{i,:}),  \label{eq:naive}
\end{gather}
where $\tau > 0$ is the temperature of the softmax function.
Note that $\mY$ was computed from the full sequence $\vx$ including the value of $x_i$, so this only approximately holds. 
This \textit{wildtype-marginal} approximation \citep{meier2021language} allows a single $O(L^2)$ forward-pass\footnote{We assume that MLMs use quadratic complexity scaled dot-product attention.}, by replacing $\hat{p}(x_i=a_r|{\vx}_{j \ne i})$ with $\hat{p}(x_i=a_r|{\vx})$.

To obtain the true conditional probability distribution $\hat{p}(x_i|{\vx}_{j \ne i})$, one instead masks $x_i$ such that $\tilde{\vx}^{(i)} = [x_1 \varlist x_{i-1}, \mathtt{MASK}, x_{i+1} \varlist x_L]$, and computes MLM output $\tilde{\mY}^{(i)}$, thus obtaining
\begin{gather}
    \hat{p}(x_i=a_r|{\vx}_{j \ne i}) = \texttt{softmax}_\tau(\tilde{\mY}^{(i)}_{i,:}).  
\end{gather}
A separate forward-pass is needed for each $i$, and each time $L-1$ rows of $\mY$ are discarded. 
Thus each iteration of mutation-centric sampling has cost $O(L^3)$. 
 
Unlike CLMs, MLMs do not offer an efficient way to compute a sequence log-likelihood or perplexity.
Instead, a pseudo-log-likelihood (PLL) is computed by summing over positions $\sum_i \log \hat{p}(x_i=a_r|{\vx}_{j \ne i})$, for a total cost of $O(L^3)$ \citep{salazar2020masked,wang2019bert}; in contrast, the CLM true log-likelihood (LL) costs only $O(L^2)$.

\subsection{Sequence-centric search via MLM pseudo-log-likelihoods}

Instead of asking the MLM to generate mutations, here we propose to ask the MLM to evaluate sequences via its PLL, which allows us to reduce our task to a search problem.
This would appear to be infeasible given the cost of computing the PLL, but it turns out that having computed the PLL for a single sequence, the (approximate) PLLs for all sequences that are one substitution away come for free.
Let $\vx'$ be a sequence that is one substitution away from template $\vx$ at position $k$.
We then use an approximation to the PLL given by 
\begin{gather}
    PLL(\vx';\tau) \approx \sum_{i=1}^{L} \log\big(\texttt{softmax}_\tau(\tilde{\mY}^{(i)}_{i,\texttt{index}(x'_i)})\big)
    \label{eq:pll-beam}
\end{gather}
where we use the exact conditional probabilities at position $k$, while using the template's conditional probabilities $\hat{p}(x_i | \mathbf{x}_{j \neq i})$ elsewhere.
This latter approximation is known as the \textit{masked-marginal approximation}, widely used for zero-shot mutation effect prediction \citep{meier2021language}, including on ProteinGym \citep{notin2023proteingym}, due to its strong performance.

The ability to quickly compute approximate PLLs motivates running beam search for $E$ steps, starting from the initial seed sequence to naturally enforce proximity.
Each time we expand a sequence from the beam, it becomes the new template sequence $\vx$, and we evaluate its PLL, thus limiting the approximation error.
With a beam of size $B$, we incur a cost of $O(BEL^3)$, obtaining $20BEL$ unique sequences.
For a realistic scenario where $E=5, L=100$ and one desires a large number of sequences pre-filtering, the $20EL \times$ speedup compared to mutation-centric sampling is enormous. 

As we will see later, this approach tends to elicit higher-quality sequences from MLMs than mutation-centric sampling.
Recall that at each beam expansion, we recompute the exact stepwise-masked conditionals of the new template, so the masked-marginal approximation error never compounds across the $E$ steps.
Moreover, while mutation-centric methods commit irreversibly to each residue based on a local conditional, beam search scores whole sequences and maintains $B$ alternatives at every step.
Beam search thus lets locally-attractive but globally-incoherent paths die out, in favor of sequences the MLM endorses globally.

In practice, we would also like to obtain a diverse set of sequences.
We adapt stochastic beam search \citep{kool2019stochastic}, which adds Gumbel noise before ranking \citep{gumbel1954statistical,maddison2014sampling} to balance sequence likelihood and diversity.
The aforementioned softmax temperature, because it scales the likelihood term but not the Gumbel term, adjusts the balance.
Note that while Kool et al. proposed perturbing log-probabilities so that ranking by perturbed scores is equivalent to sampling without replacement \citep{vieira2014gumbel}, their work considered beam search for autoregressive generation. In addition to our different approximate scores, our search setting is structurally different: theirs is rooted at the empty sequence with partial-sequence nodes and branching factor $|A|$; ours is rooted at the seed with full-sequence nodes and branching factor $(|A|-1)L$.

Our approach can be applied to CLM LLs instead of MLM PLLs, if one wants to use a protein or antibody CLM to generate sequences near to a seed sequence, rather than to generate sequences from scratch. 
However, this requires $20L$ separate LL calculations, for a net $20\times$ cost increase.

\begin{table}[t]
  \centering
  \begin{tabular}{lcccc}
  \toprule
  & \multicolumn{2}{c}{Approximation} & \multicolumn{2}{c}{Approximation + Gumbel} \\
  \cmidrule(lr){2-3} \cmidrule(lr){4-5}
  Model & Pearson & Spearman & Pearson & Spearman \\
  \midrule
  ESM2-650M & 0.964 & 0.940 & 0.735 & 0.727 \\
  AbLang2 & 0.869 & 0.869 & 0.752 & 0.729 \\
  ProteinMPNN & 0.944 & 0.934 & 0.721 & 0.724 \\
  AbMPNN & 0.927 & 0.850 & 0.726 & 0.615 \\
  \bottomrule
  \end{tabular}
  \caption{Approximate-vs-true PLL correlation over all 190 one-edit CDR-H3 variants of Trastuzumab ($\tau=1$).}
  \label{tab:pll_one_edit_h3}
\end{table}

We verified the PLL approximation empirically on Trastuzumab, finding strong agreement with the exact PLL over all 190 one-edit CDR-H3 variants across four models, shown in Table \ref{tab:pll_one_edit_h3}.
The approximation alone correlates strongly with the true PLL, while adding Gumbel noise for stochastic ranking reduces these correlations as expected.
We extended the above analysis to the larger 7 antibody panel from \citet{hie2024efficient}; the results further confirm the strength of the approximation, and are provided in Appendix \ref{app:pll-masked-marginal-empirical}.
A deeper theoretical discussion of PLL approximations in beam search is given in Appendix \ref{app:pll-approximations}.

\subsection{Multi-objective optimization (MOO) with gradient-free guidance}

Our proposed search framework treats both the MLM and optional scoring functions as black boxes, neither providing partially-masked sequences nor requesting gradients.
Any multi-objective scalarization technique is thus applicable for our framework. 
Recent works addressing protein multi-objective optimization have employed Pareto non-dominated sorting (NDS) \citep{chen2025multi,tang2025peptune,zhang2025multi}.
We consider that, as well as smooth Tchebycheff scalarization (STS) \citep{lin2024smooth}, which attempts to improve the performance of all objectives, rather than any objective as does NDS.
NDS does not admit weights over objectives; in some of our experiments, we employed a weighted extension of STS.
See Appendix \ref{subsec:app-moo} for more details. 

\section{Experiments}

\subsection{\textit{In silico} method evaluation on Trastuzumab CDR-H3 optimization}

We evaluated our proposed method using Trastuzumab's VH domain; we diversify the 10-residue CDR-H3 with $E \in \{1, \dots, 7\}$ edits and evaluate generators under ESM2-650M \citep{lin2022language}, AbLang2 \citep{olsen2024addressing}, ProteinMPNN \citep{dauparas2022robust}, and AbMPNN \citep{dreyer2023inverse}. For evaluation, we retrained a CNN classification oracle on the 367k-sequence binding dataset of \citet{chinery2024simple} (reproducing its near-perfect reported test AUROC, at $0.995$); secondarily we evaluate humanness via OASis percentile \citep{prihoda2022biophi}.
For guided generation, we also retrained a CNN classification model (dubbed ``\texttt{MasonCNN}'' hereafter) on the 34k-sequence binding dataset of \citet{mason2021optimization} (reproducing its reported 0.91 test AUROC, at 0.914).
In addition to three unguided methods (denoising, Gibbs, and our beam search), we evaluate a Gibbs with \texttt{MasonCNN} PoE guidance baseline, our beam search with \texttt{MasonCNN} STS guidance, and our beam search with equally-weighted \texttt{MasonCNN} and OASis percentile guidance.
Further experimental details are given in Appendix \ref{subsec:app-her2-details}.

\begin{figure}
    \centering
    \includegraphics[width=0.99\linewidth] {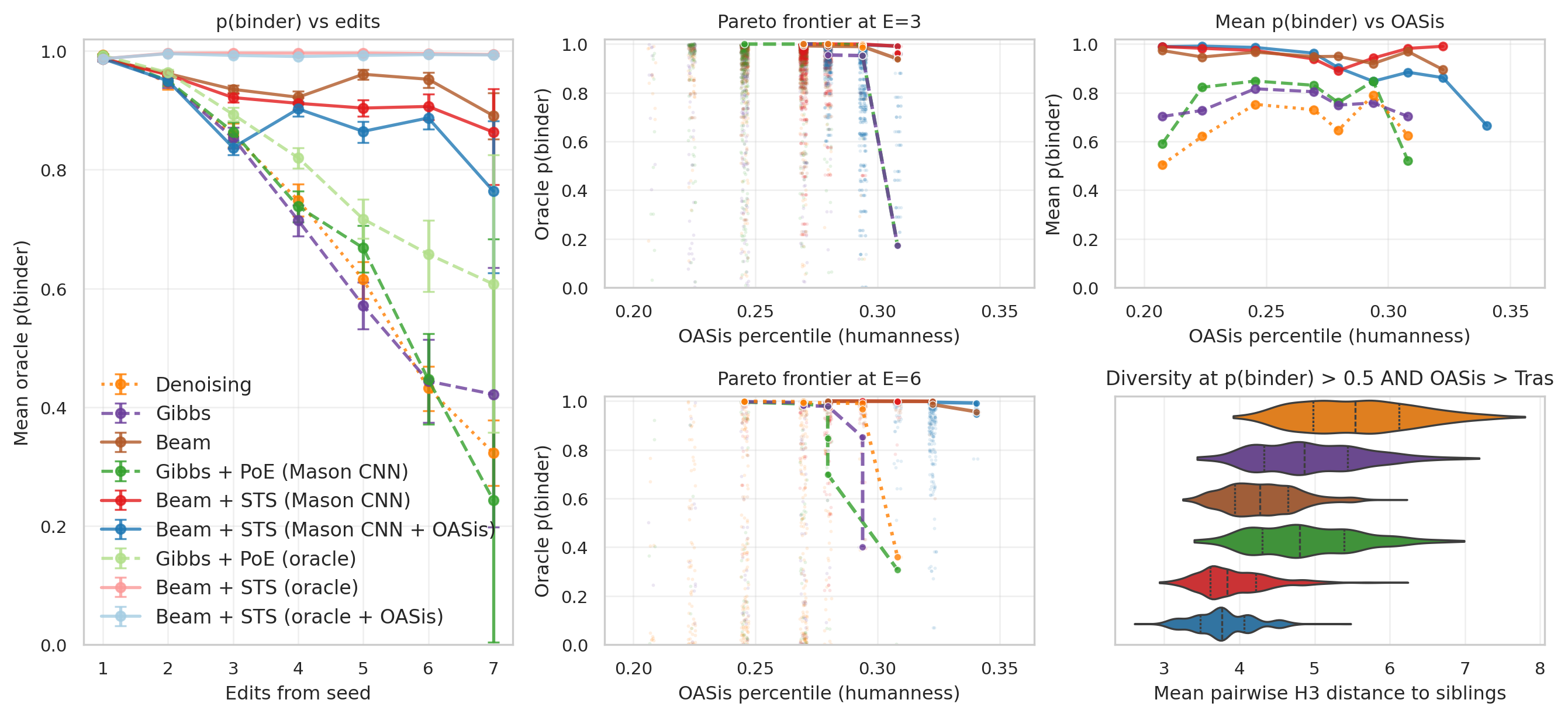}
    \caption{ESM2-650M on the Trastuzumab CDR-H3 \textit{in silico} evaluation. Color groups \texttt{MasonCNN}-guided methods with their lighter-shaded strong-oracle counterparts; linestyle indicates methodology (solid: ours; dashed: previous). \textbf{[Left:]} Mean predicted \texttt{p(binder)} vs edits from the seed; error bars depict 95\%-bootstrap CIs. \textbf{[Middle:]} OASis × \texttt{p(binder)} Pareto frontiers. Faded dots show the underlying generated variants. \textbf{[Top Right:]} Mean \texttt{p(binder)} at each OASis percentile bucket, pooling across all edits. \textbf{[Bottom right:]} Pairwise H3-diversity violins restricted to ``high-quality'' variants — those with \texttt{p(binder)} $>$ 0.5 AND OASis $>0.245$, the Trastuzumab seed.}
    \label{fig:her2-esm}
\end{figure}

Results for ESM2 are shown in Figure \ref{fig:her2-esm}.
Under ESM2, our unguided beam search dominates the mutation-centric baselines on binding (left panel): at six realized edits, mean \texttt{p(binder)} $\approx 0.95$, versus $\approx 0.43$ for denoising and $\approx 0.44$ for Gibbs. 
ESM2's antibody priors are strong enough on their own that adding \texttt{MasonCNN} guidance does not further improve binding for our approach, but pairing it with OASis humanness as a second objective reaches the Pareto-optimal upper-right region at both $E=3$ and $E=6$ (middle panel), raising humanness from the Trastuzumab-seed while preserving high \texttt{p(binder)}, and dominating every Gibbs or denoising-based method.
We observed that \texttt{MasonCNN} guidance did not improve binding for Gibbs; Gibbs PoE guidance with the strong oracle showed somewhat improved binding probability, but still worse binding than even our unguided beam search (left panel).
We observed somewhat less pairwise sequence diversity among generated variants passing basic quality filters (bottom right panel), with Beam + STS (\texttt{MasonCNN} + OASis) the most concentrated: multi-objective guidance trades a controlled amount of diversity for higher reliability.

Results for AbLang2, ProteinMPNN, and AbMPNN, are qualitatively similar, and are presented in Appendix \ref{subsec:app-her2-results}. 
Notably, we find that AbLang2 likelihoods are negatively correlated with \texttt{p(binder)}, necessitating supervised guidance; ProteinMPNN and AbMPNN have performance strongly dependent on the availability of antigen structure; and that ESM2-650M is at least as good as the inverse-folding MPNNs even with the antigen.

We also use this setting to evaluate the behavior of our approach as a function of the beam size and temperature hyperparameters, with results in Appendix \ref{subsec:app-hyperparameters}. 
These results show that quality is very robust to the choice of hyperparameters, and that beam size could have been increased to larger values than used in our experiments, increasing diversity without decreasing quality.
Furthermore, using this setting, we empirically demonstrate the substantial computational speedup from our approach, with results in Appendix TODO.

\subsection{\textit{In silico} benchmarking}

We then evaluated a wide variety of combinations of models and sampling algorithms \textit{in silico}.
We evaluated nine MLMs: ESM-2 (35M, 150M, and 650M params) \citep{lin2022language}, Sapiens (500K) \citep{prihoda2022biophi}, AbLang-2 (45M) \citep{olsen2024addressing}, AMPLIFY (120M and 350M) \citep{fournier2024protein}, DiffAbOpt \citep{raghu2025guided} (a CDR-only model that also conditions on a seed's predicted structure \citep{passaro2025boltz}), and an internally-trained 13M parameter BERT-style model \citep{devlin2019bert} trained on \citep{dunbar2014sabdab}; all but ESM-2 were trained on \citep[OAS;][]{kovaltsuk2018observed} antibody sequences.
We also evaluated three CLMs: pIgGen and pIgGen-developability (17M) \citep{turnbull2024p}, and CloneLM (377M) \citep{amin2024bayesian}, all of which were trained on antibodies.
These methods were evaluated on a real scFv antibody therapeutics program for which we have supervised oracles for predicting synthesizability, binding affinity, and thermostability (melting temperature); we also evaluated OASis percentile and isoelectric point.
Each method was given the same set of 10 seeds; we evaluated generated variant sequences with exactly 3 substitutions.

An initial evaluation of sampling algorithms on AbLang2 found that Gibbs sampling outperformed denoising in predicted synthesizability, and produced more diversity than Gibbs-argmax (Appendix Figure \ref{fig:ablang-samplers}).
We therefore focused our evaluations of MLMs with these samplers, and of CLMs with our beam search.

\begin{figure}
    \begin{tabular}{l @{\hspace{1cm}} l}
       (A) & (B) \\
       \includegraphics[width=0.4\linewidth]{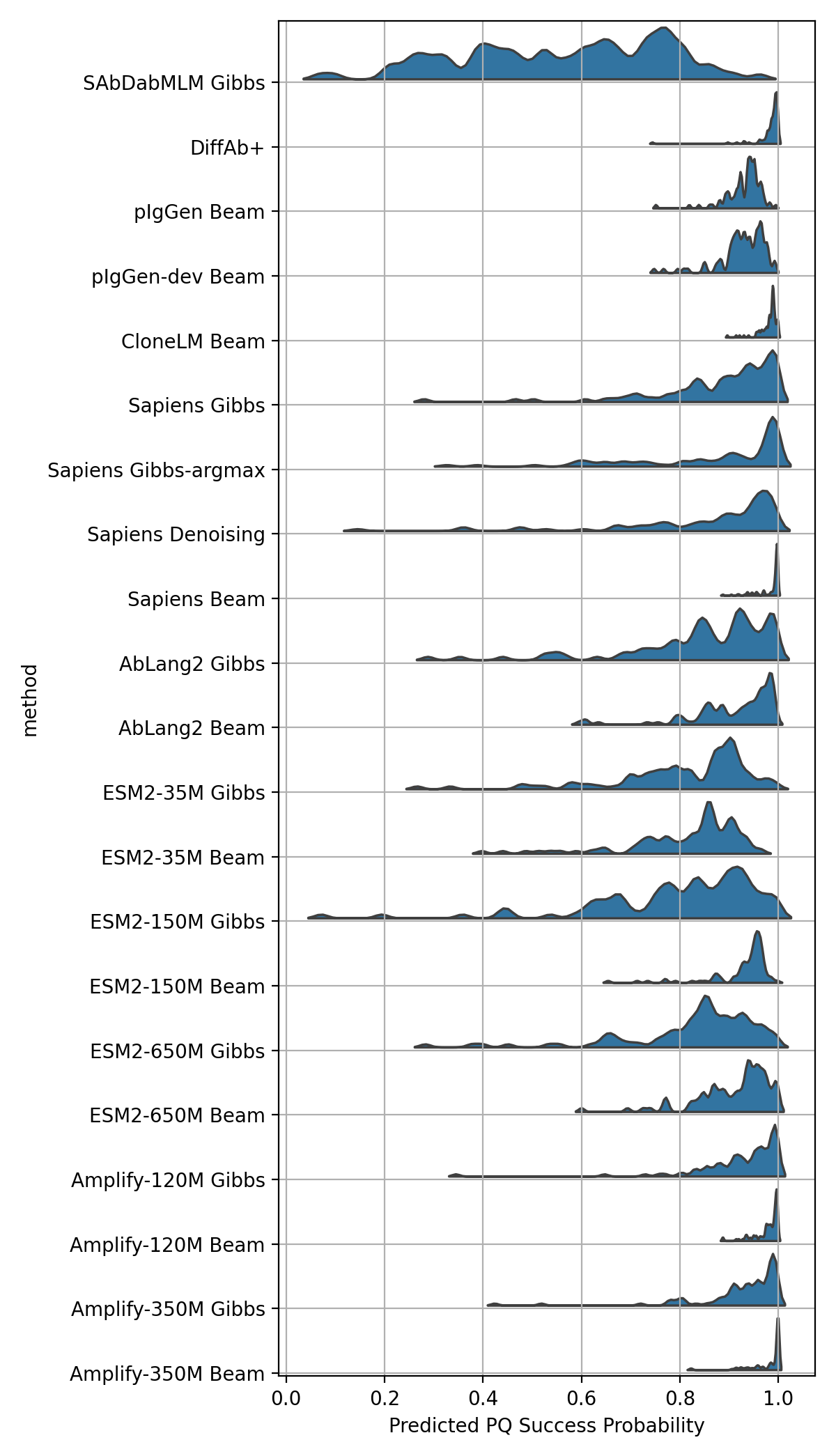}  & \includegraphics[width=0.4\linewidth]{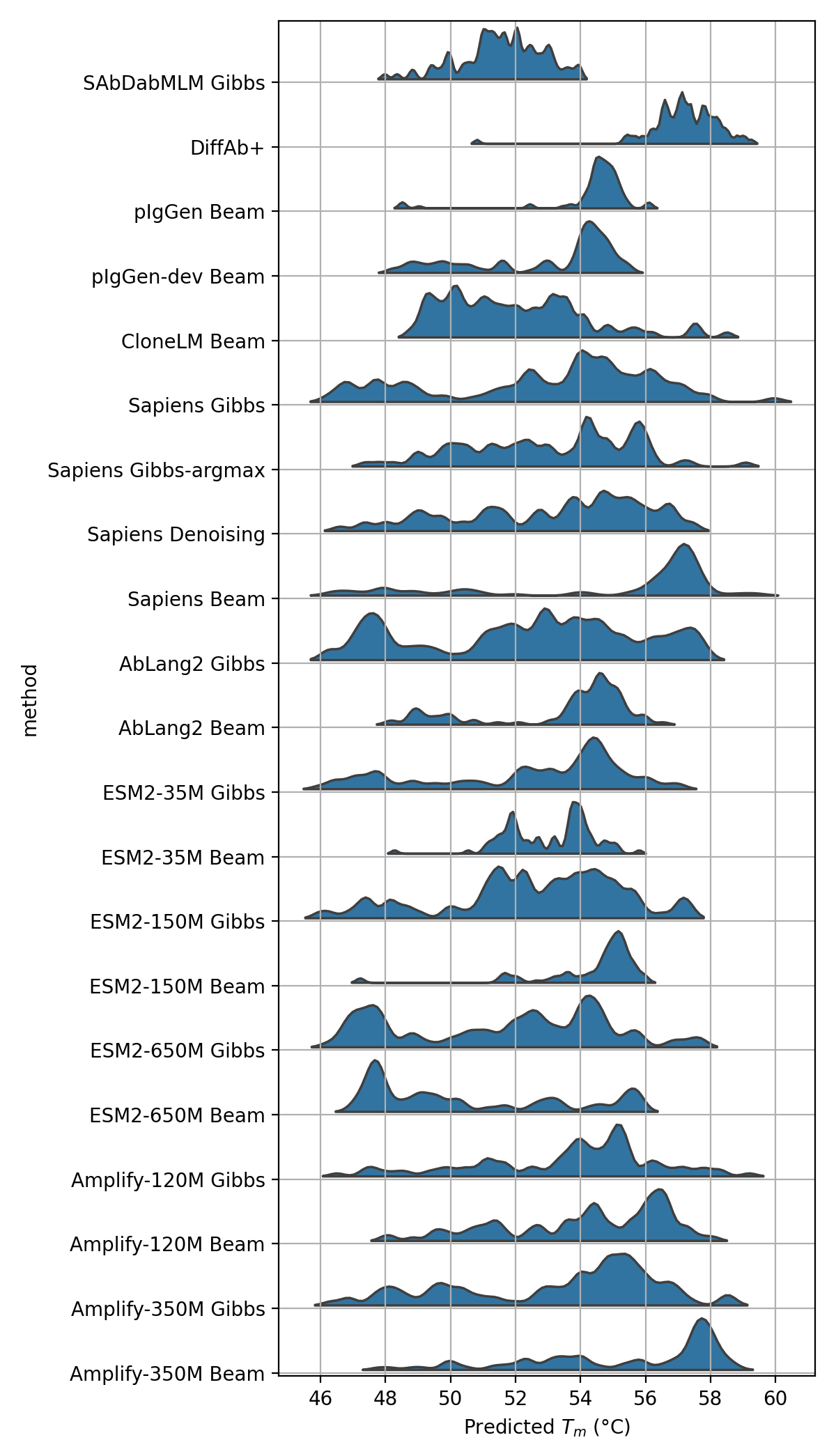} \\
       (C) & (D) \\
       \includegraphics[width=0.4\linewidth]{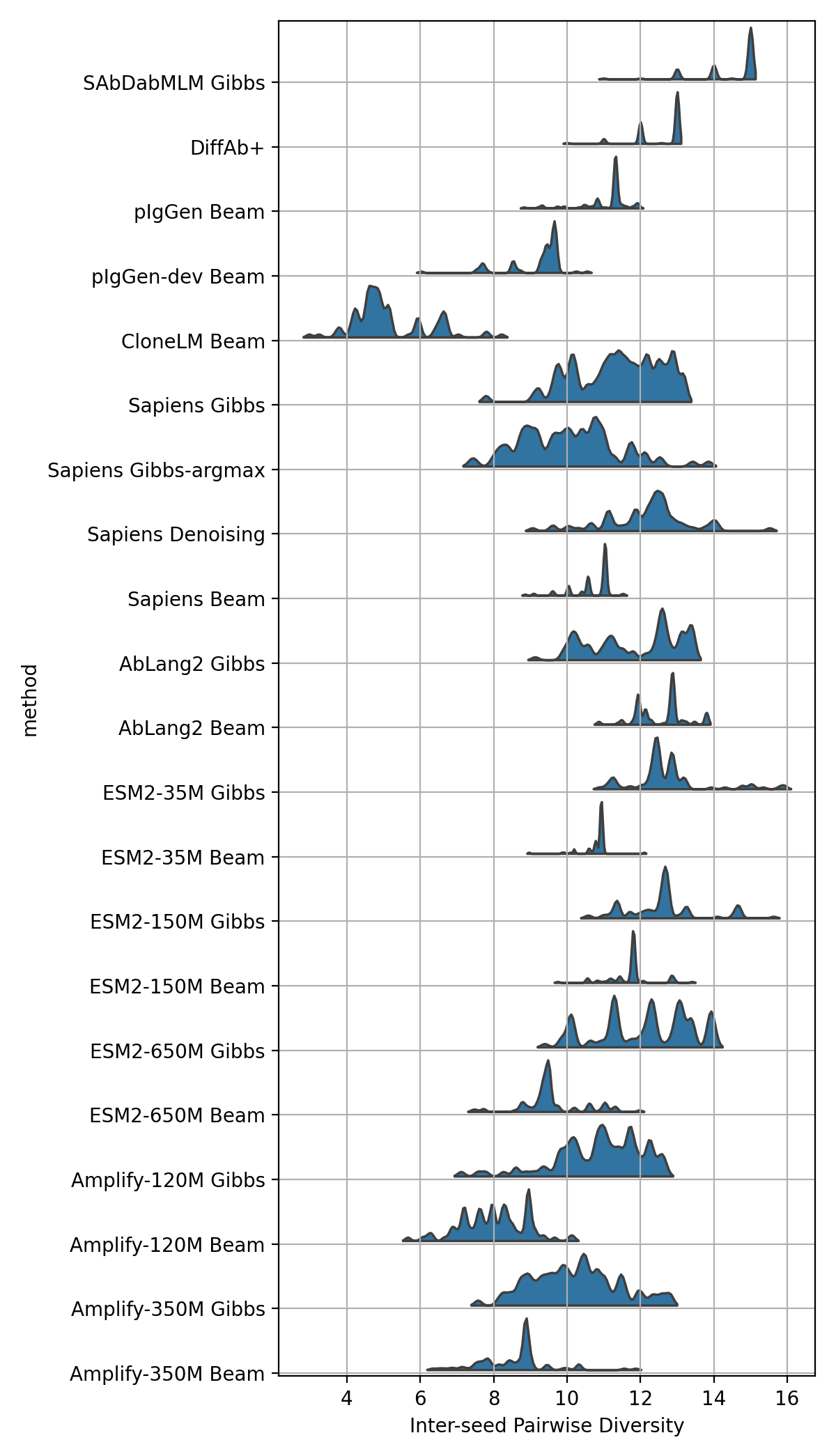}  & \includegraphics[width=0.4\linewidth]{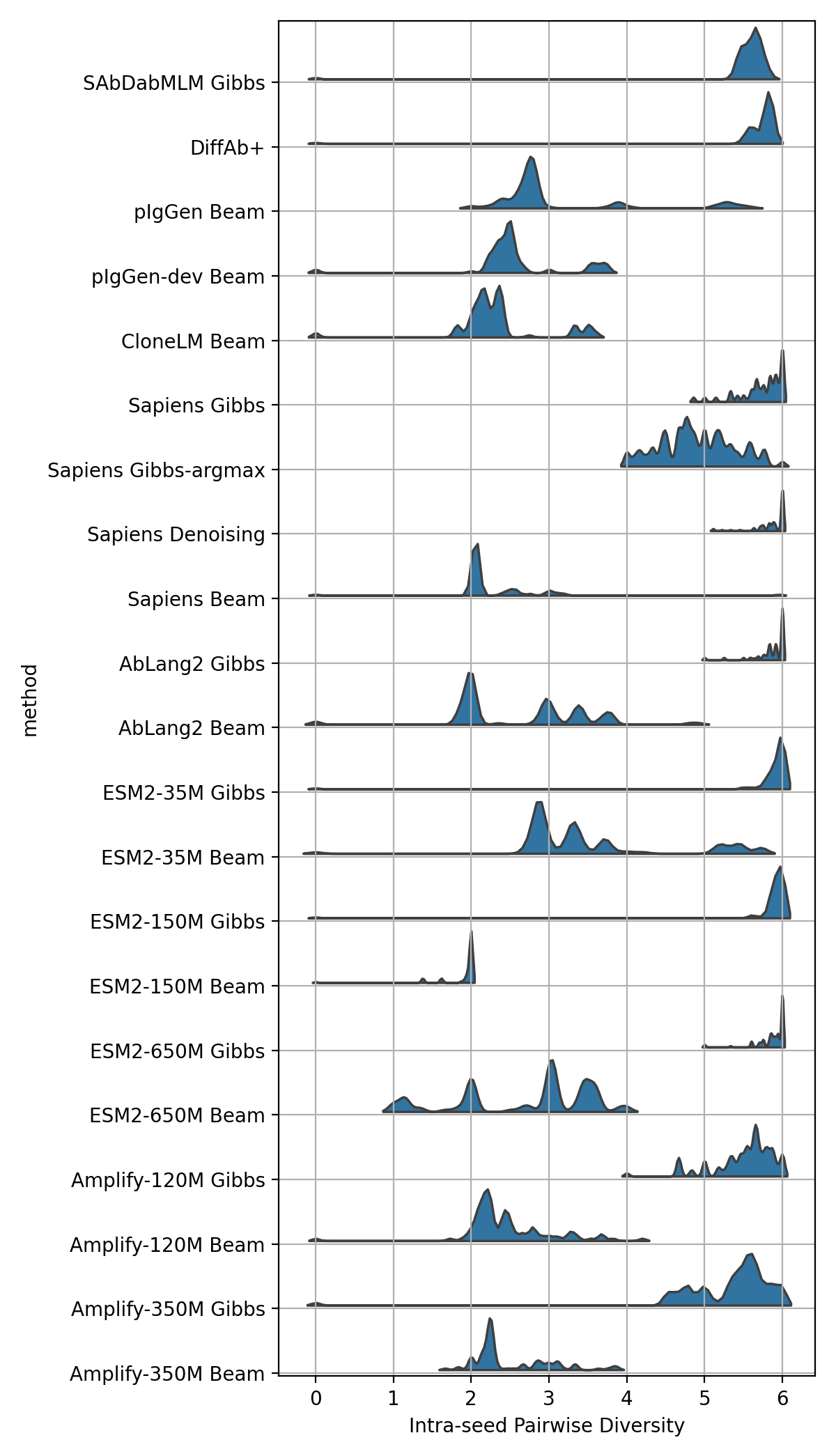} \\       
    \end{tabular}
    \centering    
    \caption{\textit{In silico} metrics of quality and diversity for generated sequences: (A) predicted probability of synthesizability, (B) predicted melting temperature, (C) inter-seed pairwise diversity, and (D) intra-seed pairwise diversity.}
    \label{fig:insilico-quality-diversity}
\end{figure}

Overall, the AbLang2 and ESM2-650M models distinguished themselves with excellent metrics across a variety of criteria, and our beam search outperformed Gibbs sampling.
We first show results for predicted synthesizability (Fig. \ref{fig:insilico-quality-diversity}A) and predicted thermostability (Fig \ref{fig:insilico-quality-diversity}B).
Methods using our proposed beam search method consistently performed better on predicted synthesizability, compared to other sampling methods;
in contrast, there was no discernable improvement for thermostability.

Pairwise diversity results are shown in Fig. \ref{fig:insilico-quality-diversity}C for inter-seed diversity and Fig \ref{fig:insilico-quality-diversity}D for intra-seed diversity.
We see that beam search has much less intra-seed diversity than Gibbs, but this pattern is much less strong for inter-seed diversity.
More generally, we see that intra-seed and inter-seed diversity are not necessarily correlated; CloneLM Beam and Amplify Beam have decent intra-seed diversity but low inter-seed diversity.
Low intra-seed diversity means that children of the same seed tend to share the same mutated positions and especially the same new residues, whereas low inter-seed diversity means that variants from different seeds are converging to the same part of sequence space; the latter is more problematic.

See Appendix \ref{subsec:app-insilico} for further experimental details and results (on humanness, germline bias, isoelectric point, and CDR mutation count).

\subsection{\textit{In vitro} benchmarking}

When choosing a smaller set of models for \textit{in vitro} evaluation, we brought forward AbLang2 and ESM2-650M for their \textit{in silico} performance.
We also evaluated Sapiens as a small-parameter-count baseline model, pIgGen-dev as a representative of CLMs, SAbDabMLM due to its non-OAS training data, and DiffAb+ to represent structure-based methods.
We applied these models to a fresh FAb antibody therapeutic campaign, evaluating chiefly on their ability to pass synthesizability and binding QC criteria.
We used an initial candidate hit as seed; for each method 2/3 of the sequences sent to the wet lab had exactly 4 CDR-only substitutions, and 1/3 had exactly 3 substitutions in any location.
(Since DiffAb+ is a CDR-only method, all its variants had exactly 4 CDR-only substitutions.)

Later, after obtaining 729 samples and training a supervised classification model for synthesizability and binding success, we evaluated the effect of combining this with AbLang2 when proposing batches.
We use the model for post-MLM filtering ($p_{success}>0.6$), post-MLM ranking, and finally NDS and STS guidance with post-MLM ranking.

\begin{figure}[H]
    \centering   \includegraphics[width=1\linewidth]{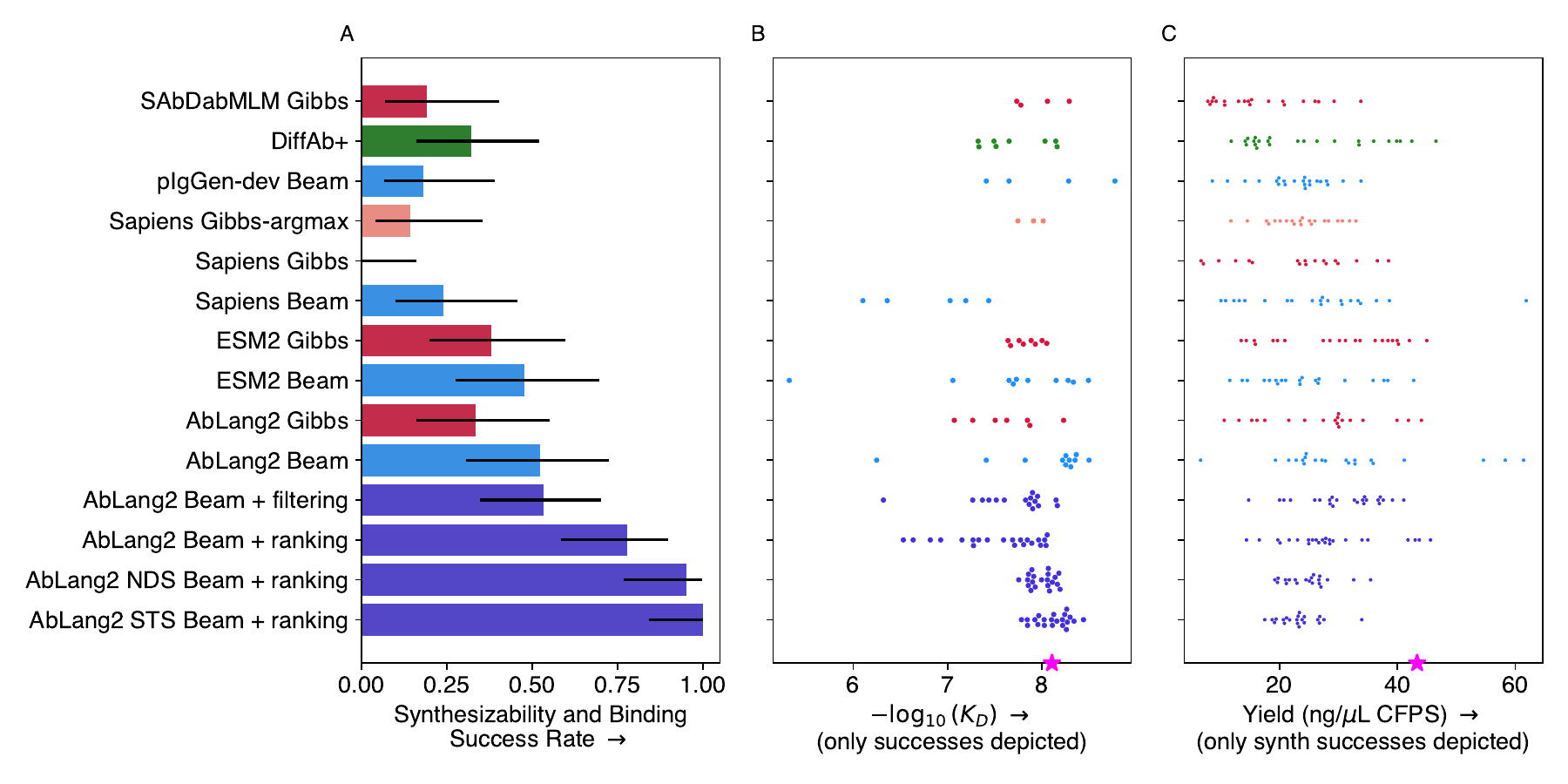}
    \vspace{-5pt}
    \caption{\textit{In vitro} results. For error bars around the success rate, we show 95\% CIs via binomial test inversion. We depict the seed's binding affinity and yield as a star on the x-axis.}
    \label{fig:invitro-main}
    \vspace{-5pt}
\end{figure}

Results, comprising 289 samples across 14 methods, each with at least 21 samples, are shown in Figure \ref{fig:invitro-main}.
The overall success rate of each method is shown in Fig \ref{fig:invitro-main}A.
Consistent with \textit{in silico} results, ESM2-650M and AbLang2 had the best unsupervised success rates.
Also, our proposed beam search outperformed Gibbs on all three models where both were tried.
On Sapiens, Gibbs-argmax outperformed Gibbs, but it struggled to propose many unique new sequences.
We found that using supervision to rank the unguided AbLang2 outputs improved success rate considerably.
Using NDS and STS MOO guidance during generation improved results even further, with STS MOO guidance leading to a perfect 100\% success rate.

In Fig \ref{fig:invitro-main}B, we see that methods with higher success rate usually had tighter binding on their successful antibodies.
Notable exceptions were SAbDabMLM and pIgGen-dev, perhaps due to their idiosyncratic training data.
Both NDS and STS guidance not only improved success rate, but eliminated generation of very weak binders.
In Fig \ref{fig:invitro-main}C, for yield relative to CFPS input volume (a quantitative measure of synthesizability), we do not see any substantial trends among unguided methods.
We do see, though, that our guidance methods produced antibodies with less variance in yield, particularly on the higher end of the distribution.

\begin{figure}[H]
    \centering   \includegraphics[width=1\linewidth]{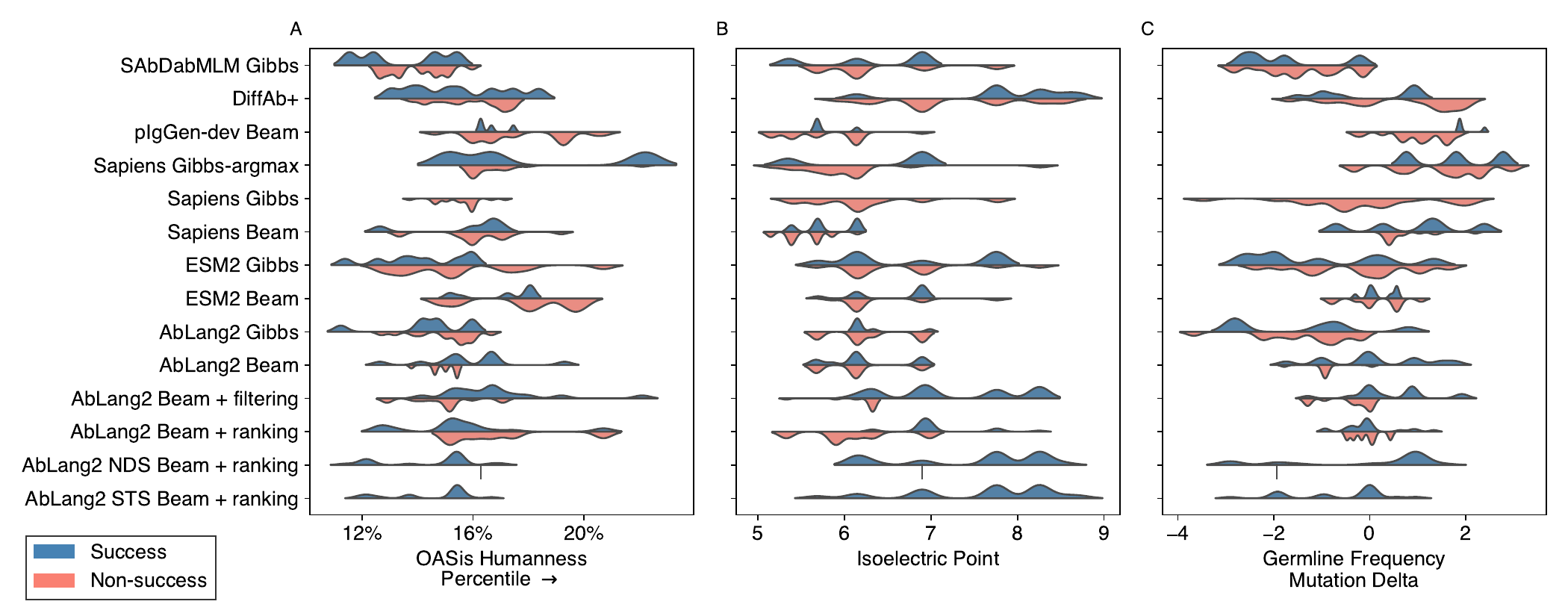}
    \vspace{-5pt}
    \caption{Additional developability-related \textit{in silico} evaluations of the sequences in the \textit{in vitro} experiment. For each method, observations are split by successful synthesizability and binding.}
    \label{fig:invitro-developability}
    \vspace{-5pt}
\end{figure}

\begin{figure}[H]
    \centering   \includegraphics[width=1\linewidth]{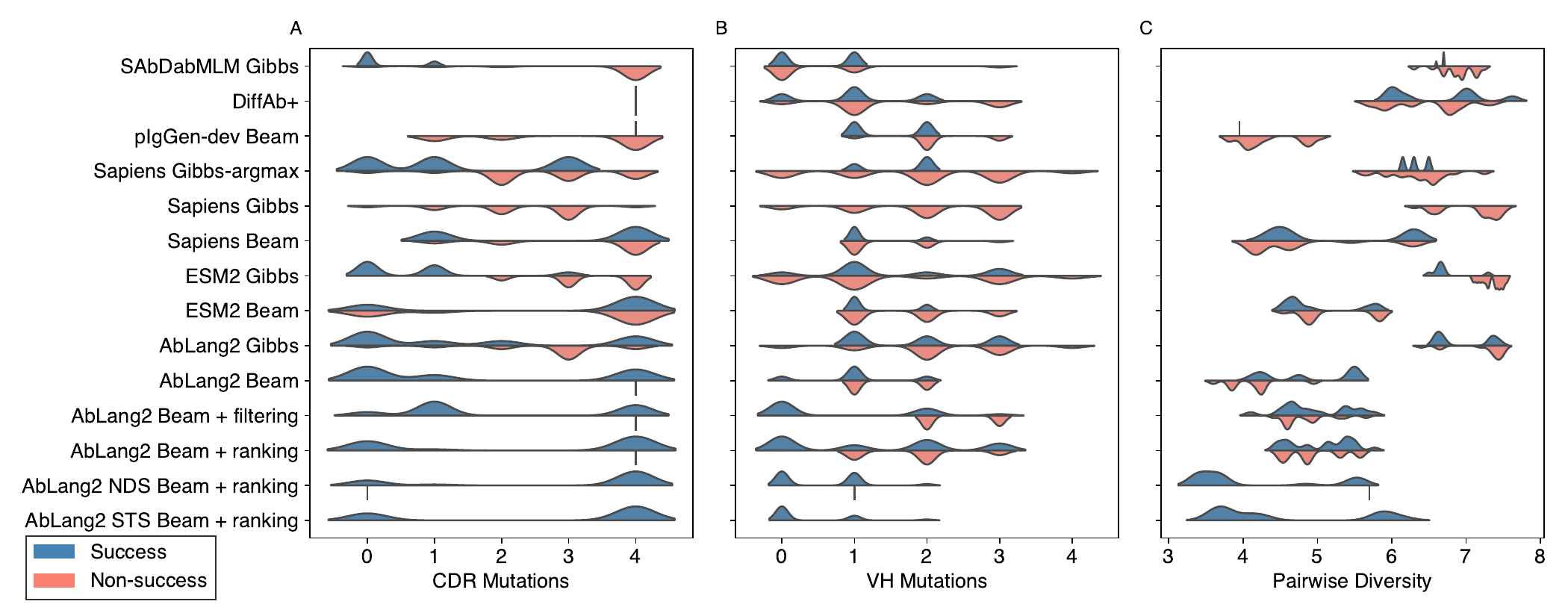}
    \vspace{-5pt}
    \caption{Additional analysis of mutational preferences of different methods for the \textit{in vitro} experiment. or each method, observations are split by successful synthesizability and binding.}
    \label{fig:invitro-mutations}
    \vspace{-5pt}
\end{figure}

We show additional \textit{in silico} evaluations of the \textit{in vitro} experiment sequences in Fig \ref{fig:invitro-developability} and Fig \ref{fig:invitro-mutations}.
In Fig \ref{fig:invitro-developability}A, showing OASis humanness, we once again witness a pattern of superiority for our proposed beam search methods over their Gibbs counterparts.
On the other hand, we surprisingly see that AbLang2 tended to produce less human antibodies, despite being trained on human sequences; meanwhile, ESM2 650M had fairly high humanness outputs, despite not even being trained on antibody-specific data.
SAbDabMLM sequences have low humanness, unsurprisingly since it was not trained on OAS.
We do see that the guidance methods produced less human antibodies, a risk of using a single guidance model which can in principle be ameliorated by guidance with a humanness metric.


In Fig \ref{fig:invitro-developability}C, we show germline bias. 
Positive values indicate mutations toward more germline-typical residues. 
Recall that while increased humanness is usually good for antibody therapeutics optimization (i.e. reduced immunogenicity), positive germline bias can be problematic for binding specificity \citep{olsen2024addressing}.
We see that Sapiens Gibbs-argmax and Sapiens Beam, but not Sapiens Gibbs, have high germline bias.
This is consistent with the observation in \citet{olsen2024addressing} that Sapiens has high germline bias, and with our observation that Gibbs sampling often fails to elicit sequences that are consistent with that model's preferences.

In Fig \ref{fig:invitro-mutations}A, we reassuringly see that the successful supervised methods still heavily mutated the CDR regions, like the unsupervised methods.
On the other hand, in Fig \ref{fig:invitro-mutations}B, we see that they were systematically different from unsupervised methods in that they more frequently mutated the VL rather than the VH chain.
In Fig \ref{fig:invitro-mutations}C, showing pairwise diversity, we see that our beam search method and especially our guidance-based methods had less pairwise diversity.
However, it's important to note that this experiment reflects diversity when using a single seed.
As shown in our \textit{in silico} evaluation, beam search had less intra-seed diversity, while maintaining inter-seed diversity.

\section{Conclusion}
We proposed a new MLM sampling method for iterative protein optimization, departing from the dominant mutation-centric paradigm and replacing it with sequence-centric evaluation and search.
We benchmarked sampling methods and several language models on real antibody therapeutics optimization campaigns.
Our results yield new insights into the effects of models, sampling algorithms, and guidance.

We conclude with several practical recommendations. 
First, we advocate for the use of supervision, for both ranking and guidance, whenever labeled data may be acquired.
Second, we are able to recommend ESM2-650M and especially AbLang2 for antibody engineering.
Third, we advise using our proposed stochastic beam search over Gibbs-type sampling. 
Fourth, we highlight the need for care when using supervised guidance, as it can lead to undesirable side effects in multi-objective optimization settings.
Fifth, we suggest examining smooth Tchebycheff scalarization as an alternative to non-dominated Pareto-based sorting, for settings where one wants to simultaneously maximally-satisfy multiple objectives, rather than make progress along each objective.

\newpage
\bibliography{output}
\bibliographystyle{plainnat}

\newpage
\appendix
\section{Appendix}

\subsection{An empirical evaluation of the masked-marginal approximation}
\label{app:pll-masked-marginal-empirical}

We performed a deeper dive into the validity of our approximation for a larger set of seed sequences, for paired-chain vs unpaired-chain modeling, and for CDR-region-only vs CDR+framework full-sequence. The results, reinforce the validity of the approximation, and are also interesting in their own right. In summary, we extended the empirical analysis from our earlier rebuttal, and almost always observed the same strong correlation between approximate and true PLLs.

The exception is an insightful corner case: it struggles only near-germline seeds, intersected with framework mutations, intersected with AbLang2. Digging deeper, we found strong evidence that this is aberrant behavior with AbLang2, where its ``true'' PLL landscape has non-biological anomalies from memorizing germline framework sequences, as opposed to true biological epistasis. (One would avoid using an LM with such behavior on such regions and seeds, so the weak correlation is a moot point. And our approximate-PLL was indeed better biologically than the ``true-PLL'' when they disagreed.)

Our main findings, on the 7 antibody panel from \citet{hie2024efficient}, showing the approximate-vs-true PLL Spearman correlations, on full (framework and CDR) VH-plus-VL sequences, are shown in Table \ref{tab:pll_one_edit_hie_by_model}. We see strong approximate-vs-true Spearman correlations for Sapiens (antibody LM) and ESM2-650M, and even for AbLang2 except for antibodies \texttt{medi8852-uca} and \texttt{mab114-uca}. Note the "-uca" (unmutated common ancestor, UCA): these are unobserved but inferred consensus germline sequences. Because UCAs are inferred germline ancestors with no experimental structures, inverse-folding MPNNs were skipped.

\begin{table}[h!]
\centering
\begin{tabular}{lccc}
\toprule
Antibody & Sapiens & AbLang2 & ESM2-650M \\
\midrule
\texttt{c143} & 0.941 & 0.955 & 0.930 \\
\texttt{mab114} & 0.924 & 0.935 & 0.896 \\
\texttt{mab114-uca} & 0.946 & 0.734 & 0.928 \\
\texttt{medi8852} & 0.837 & 0.925 & 0.878 \\
\texttt{medi8852-uca} & 0.878 & 0.282 & 0.888 \\
\texttt{regn10987} & 0.910 & 0.933 & 0.942 \\
\texttt{s309} & 0.879 & 0.854 & 0.929 \\
\bottomrule
\end{tabular}
\caption{Approximate-vs-true PLL Spearman correlation, one-edit full VH+VL variants, over the seven-antibody panel from \citet{hie2024efficient}.}
\label{tab:pll_one_edit_hie_by_model}
\end{table}

We will now provide evidence that AbLang2's ``true'' PLL landscape is degenerate on UCA seeds due to germline framework memorization. First, we isolate the weak correlation on these seeds to the frameworks. Second, we show that AbLang2 has extremely low perplexity on these seeds, surrounded by landscape cliffs, indicating memorization. Finally, we show that DASM \citep{matsen2026separating}, a new model that explicitly controls for ``germline bias'' in a principled manner, has better correlation with the approximate PLLs than the true PLLs, precisely on these seeds with AbLang2. In other words, near the germline where the approximate-AbLang2 and true-AbLang2 disagree, the true-AbLang2 is the \textit{odd one out} compared to approximate-AbLang2, DASM, and the other MLMs' PLLs.

In this paragraph, we provide some additional background on antibody ``germline bias'', which may be useful to some readers for the subsequent analysis.
The human body does not store a catalog of finished antibodies, but rather a few hundred inherited gene fragments (the ``germline'' segments), which get shuffled and then locally mutated to produce each mature antibody. Most of the sequence (the ``framework'') stays close to the inherited template, while the mutations concentrate in three sequence regions (the CDRs) that actually touch the antigen that is being targeted by the immune system. Because every LM training sequence is therefore a lightly-edited copy of one of those few hundred templates, an antibody LM can score well on masked-token recovery by filling in the germline amino acids. But such a model then proposes germline reversions, assigning low likelihood to exactly the mutations that confer binding to specific antigens.
AbLang2 was the first LM that attempted to avoid germline bias, via ML-based heuristics. We hypothesize that AbLang2 avoided having its entire likelihood landscape sloped towards germlines, by instead concentrating probability mass onto the immediate neighborhood of germline sequences (thus making them extremely high germline bias), while being relatively flat and low germline bias below these ``cliffs''. DASM is a newer model that seeks to directly model the above process to eliminate germline bias, though it is not an LM, providing a relative ``selection factor'' given a (sequence, mutation)-tuple, rather than a PLL given a sequence.

Now we provide multiple independent pieces of evidence that the ``true'' AbLang2 PLLs on these UCAs are erroneous and degenerate.

First, we isolate that the disagreement between ``true'' and masked-marginals comes from the inclusion of framework region mutations. We exclude framework mutations, including only CDR region (Chothia definition) positions into the PLL; we still condition on the frameworks by including the full sequence in the forward pass. As shown in Table \ref{tab:pll_one_edit_hie_cdr_by_model}, this mostly rescues the UCA correlations on AbLang2, while leaving others basically untouched.

\begin{table}[h!]
\centering
\begin{tabular}{lccc}
\toprule
Antibody & Sapiens-CDR & AbLang2-CDR & ESM2-CDR \\
\midrule
\texttt{c143} & 0.971 & 0.930 & 0.892 \\
\texttt{mab114} & 0.933 & 0.859 & 0.880 \\
\texttt{mab114-uca} & 0.946 & 0.855 & 0.946 \\
\texttt{medi8852} & 0.828 & 0.881 & 0.906 \\
\texttt{medi8852-uca} & 0.778 & 0.675 & 0.895 \\
\texttt{regn10987} & 0.887 & 0.927 & 0.919 \\
\texttt{s309} & 0.898 & 0.831 & 0.917 \\
\bottomrule
\end{tabular}
\caption{Approximate-vs-true PLL Spearman correlation over the seven-antibody panel from Hie et al., scoring only Chothia-defined CDR positions of the one-edit variant (while still conditioning on the full VH+VL sequence in the forward pass).}
\label{tab:pll_one_edit_hie_cdr_by_model}
\end{table}

Second, we show that AbLang2 uniquely gives extremely large likelihoods specifically to the UCA seeds, revealing memorization. In Table \ref{tab:seed_perplexity_hie}, we show per-residue pseudo-perplexity of the full seed sequences. The UCA seeds have unusually low perplexity, particularly for the antibody LMs.

\begin{table}[h!]
\centering
\begin{tabular}{lccc}
\toprule
Antibody & Sapiens & AbLang2 & ESM2 \\
\midrule
\texttt{c143} & 1.878 & 1.955 & 3.301 \\
\texttt{mab114} & 2.107 & 1.949 & 3.967 \\
\texttt{mab114-uca} & 1.256 & 1.256 & 2.862 \\
\texttt{medi8852} & 1.607 & 1.761 & 3.719 \\
\texttt{medi8852-uca} & 1.212 & 1.156 & 3.653 \\
\texttt{regn10987} & 1.407 & 1.447 & 2.838 \\
\texttt{s309} & 1.559 & 1.640 & 2.834 \\
\bottomrule
\end{tabular}
\caption{Per-residue pseudo-perplexity of the full seed sequence, over the seven-antibody panel from Hie et al.
AbLang2 uniquely assigns extremely low perplexity to the UCA seeds, revealing memorization.}
\label{tab:seed_perplexity_hie}
\end{table}

Third, we observe that AbLang2's UCA likelihood peaks are indeed surrounded by sharp cliffs in the local landscape. Table \ref{tab:neighbor_perplexity_ratio_hie} shows the geometric mean pseudoperplexity ratios for the seeds, confirming the cliffs around the AbLang2 UCAs.

\begin{table}[h!]
\centering
\begin{tabular}{lccc}
\toprule
Antibody & Sapiens & AbLang2 & ESM2 \\
\midrule
\texttt{c143} & 88.4 & 1123.0 & 124.0 \\
\texttt{mab114} & 70.6 & 1141.9 & 74.8 \\
\texttt{mab114-uca} & 175.8 & 6837.3 & 175.1 \\
\texttt{medi8852} & 119.3 & 1420.3 & 87.0 \\
\texttt{medi8852-uca} & 173.3 & 8668.2 & 94.4 \\
\texttt{regn10987} & 150.1 & 3189.8 & 191.5 \\
\texttt{s309} & 119.5 & 1826.4 & 238.2 \\
\bottomrule
\end{tabular}
\caption{Geometric-mean pseudo-perplexity ratio of the seed's local one-edit neighborhood (neighbor residue vs. seed residue), over the seven-antibody panel from Hie et al. AbLang2's UCA likelihood peaks are surrounded by sharp cliffs in the local landscape.}
\label{tab:neighbor_perplexity_ratio_hie}
\end{table}

Finally, and most convincingly, we compare the PLLs against DASM's germline bias-free selection factors. We compute DASM selection factors for the seed and each variant's mutation; then we measure the Spearmans between approximate PLLs vs DASM, and between true vs DASM. In Table \ref{tab:lm_vs_dasm_hie}, on the exact (UCA, AbLang2) intersection where approximate-vs-true correlation is weak, we see that DASM agrees more with ``approximate'' rather than ``true''. Indeed, on these seeds, DASM is negatively correlated with the AbLang2 ``true'' PPLs! This confirms that the AbLang2 true PLL landscapes around UCAs reflect degeneracy driven by germline bias, not biological selection lost by the masked-marginal approximation.

\begin{table}[h!]
\centering
\begin{tabular}{lcccccc}
\toprule
& \multicolumn{2}{c}{Sapiens} & \multicolumn{2}{c}{AbLang2} & \multicolumn{2}{c}{ESM2} \\
\cmidrule(lr){2-3} \cmidrule(lr){4-5} \cmidrule(lr){6-7}
Antibody & Approx & True & Approx & True & Approx & True \\
\midrule
\texttt{c143} & 0.34 & 0.35 & 0.53 & 0.49 & 0.53 & 0.52 \\
\texttt{mab114} & 0.39 & 0.39 & 0.52 & 0.43 & 0.55 & 0.53 \\
\texttt{mab114-uca} & 0.08 & 0.08 & 0.33 & -0.09 & 0.57 & 0.55 \\
\texttt{medi8852} & 0.24 & 0.19 & 0.45 & 0.30 & 0.52 & 0.51 \\
\texttt{medi8852-uca} & 0.11 & -0.01 & 0.28 & -0.35 & 0.54 & 0.52 \\
\texttt{regn10987} & 0.23 & 0.23 & 0.39 & 0.22 & 0.52 & 0.51 \\
\texttt{s309} & 0.25 & 0.27 & 0.44 & 0.16 & 0.54 & 0.54 \\
\bottomrule
\end{tabular}
\caption{Spearman correlation of each model's approximate ($\approx$) and true PLL against the DASM
germline-bias-free selection score, per antibody, over the seven-antibody panel from \citet{hie2024efficient}. On the UCA seeds where AbLang2's approximate-vs-true correlation is weakest, DASM agrees more with the approximation than with the true PLL -- and is negatively correlated with the AbLang2 true PLL, indicating the ``true'' landscape reflects germline bias rather than biological selection.}
\label{tab:lm_vs_dasm_hie}
\end{table}

\subsection{Approximations to the pseudo-log-likelihood in beam search}
\label{app:pll-approximations}                                                          
Recall that our beam search scores each candidate sequence $\vx'$ (which differs from the template $\vx$ by a substitution at position $k$) using an approximate PLL (Eq.~\ref{eq:pll-beam}). 
Here we discuss the space of approximations and their computational tradeoffs.

The exact PLL of $\vx'$ is 
\begin{gather}
  PLL(\vx') = \sum_{i=1}^{L} \log \hat{p}(x'_i \mid \vx'_{j \ne i}),
\end{gather}
where each term requires masking position $i$ in the \emph{child} sequence $\vx'$ and running a forward pass. 
Since $\vx'$ differs from $\vx$ at position $k$, the conditional distributions $\hat{p}(\cdot \mid \vx'_{j \ne i})$ differ from those of the template $\hat{p}(\cdot \mid \vx_{j \ne i})$ at every position $i \ne k$ (because the context includes $x'_k$ instead of $x_k$). 
Computing the exact PLL therefore requires $L$ forward passes \emph{per child}.
With $O(L)$ children per beam search step (one per position per amino acid), the total cost is $O(L^4)$.

The key insight behind our beam search is that having computed the template's PLL -- which requires $L$ forward passes yielding the matrix $\mA$ where $\mA_{i,:} = \texttt{softmax}_\tau(\tilde{\mY}^{(i)}_{i,:})$ -- one can cheaply approximate the PLL of all single-substitution neighbors. 
The question is what approximation to use at positions $i \ne k$.

Table~\ref{tab:pll-approximations} summarizes the five approaches.
They are distinguished by what the model observes at positions $i$ and $k$ when computing the conditional probability at position $i \ne k$, as well as their computational cost.
We note that position $k$ always uses the exact single-mask conditional $\mA_{k, \texttt{index}(x'_k)}$ in all approaches except the no-masking variants.

\begin{table}[h]
\centering
\caption{Comparison of PLL approximations for scoring all single-substitution neighbors during one step of beam search. Position $k$ is the substitution site; position $i \ne k$ is any other position. Each forward pass has cost $O(L^2)$.}
\label{tab:pll-approximations}
\begin{tabular}{lccl}
\toprule
\textbf{Approximation} & \textbf{Pos.\ $i$ sees at $i$} & \textbf{Pos.\ $i$ sees at $k$} & \textbf{Total cost} \\
\midrule
Exact PLL & \texttt{MASK} & $x'_k$ (child) & $O(L^4)$ \\
Double-mask & \texttt{MASK} & \texttt{MASK} & $O(L^4)$ \\
Masked-marginal & \texttt{MASK} & $x_k$ (template) & $O(L^3)$ \\
No-masking (child) & $x'_i$ (visible) & $x'_k$ (child) & $O(L^3)$ \\
No-masking (template) & $x_i$ (visible) & $x_k$ (template) & $O(L^2)$ \\
\bottomrule
\end{tabular}
\end{table}

\paragraph{Masked-marginal (our approach).}
At position $i \ne k$, we use the template's own conditional distribution, i.e., the distribution obtained by masking position $i$ in $\vx$:
\begin{gather}
  PLL_{\text{wt}}(\vx';\tau) = \sum_{i=1}^{L} \log \mA_{i,\texttt{index}(x'_i)}.
\end{gather}
Since $x'_i = x_i$ for $i \ne k$, only the term at $i = k$ differs from the template's PLL, so all neighbor PLLs can be read off from $\mA$ with no additional forward passes.
The approximation error is that position $i$ is scored conditioned on seeing the \emph{template} token $x_k$ at position $k$, rather than the child's substituted token $x'_k$. 
The total cost is $O(L^3)$ (for the $L$ stepwise-masked forward passes to compute $\mA$), amortized over all $O(L)$ neighbors.

\paragraph{Double-mask.}
To avoid conditioning on the wrong token at position $k$, one can instead mask \emph{both} positions $i$ and $k$ simultaneously. 
Let $\tilde{\mY}^{(i,k)}$ denote the MLM output when both positions $i$ and $k$ are masked. 
The approximate PLL becomes
\begin{gather}
  PLL_{\text{dm}}(\vx';\tau) = \log \mA_{k,\texttt{index}(x'k)} + \sum_{i \ne k} \log \texttt{softmax}_\tau(\tilde{\mY}^{(i,k)}_{i,\texttt{index}(x_i)}).
\end{gather}
At position $k$ we use the same single-mask conditional as the masked-marginal; at positions $i \ne k$, we are agnostic about the substitution site. 
Computing all pairwise double-mask outputs requires $\binom{L}{2}$ forward passes. 
Together with $L$ single-mask forward passes, the total cost is $O(L^4)$.
This is as expensive as the exact PLL, while introducing approximation error. 

\paragraph{No-masking.}
At the other extreme, one can skip masking at position $i$ entirely and use $\hat{p}(x_i \mid \vx)$ as an approximation to $\hat{p}(x_i \mid \vx_{j \ne i})$ \citep{prihoda2022biophi}. 
This can be applied to either the template sequence (one forward pass shared by all neighbors, $O(L^2)$ total) or each child sequence individually ($O(L^3)$ total, since each child requires its own forward pass). 
In either case, the model observes $x_i$ itself when predicting position $i$, yielding probabilities near $1$ that contribute little discriminative signal.
At position $k$, the child no-masking variant at least conditions on the correct token $x'_k$ (just without masking it), while the template no-masking variant conditions on the wrong token $x_k$ without masking.

\paragraph{Discussion.}
The approaches are ordered in Table~\ref{tab:pll-approximations} from most to least principled. 
The exact PLL and the double-mask approach have the same asymptotic cost but differ in that the exact PLL conditions each position on the child's actual token at $k$, while the double-mask marginalizes over $k$ via the mask token. 
In practice, the masked-marginal offers an appealing balance: it is the cheapest approach that still masks position $i$ (yielding meaningful conditional probabilities), while the error from conditioning on the template token at $k$ is empirically small in practice \citep{meier2021language}.

\subsection{Multi-objective optimization guidance}
\label{subsec:app-moo}

For multi-objective optimization, we begin by converting the pseudo-log-likelihood scores to pseudo-perplexity scores.
(Each sequence's pseudo-log-likelihood was already perturbed with independent Gumbel noise $g \sim \text{Gumbel}(0,1)$ prior to computing the pseudo-perplexity.)
Then, for each objective, we standardize scores to $z$-scores using the mean and standard deviation computed across all candidate neighbors of all template sequences at the current number of edits. 
To produce a single ranking, we aggregate the $z$-scored objectives using a smooth Tchebycheff scalarization \citep{lin2024smooth}, computing $\log \sum_i \exp(c_i \tilde{f}_i)$ for each sequence, where $\tilde{f}_i$ are the $z$-scored objectives and $c_i$ are per-objective weights, with larger-magnitude $c_i$ placing greater emphasis on objective $i$; we use $c_i < 0$ for criteria which we seek to minimize.
When computing the $z$-scores, we added $\epsilon = 0.01$ to the denominator for numerical stability.
Sequences are then ranked by this score, with lower values preferred.

\subsection{Additional experimental details for Trastuzumab optimization}
\label{subsec:app-her2-details}

\paragraph{Trastuzumab} The seed is Trastuzumab's variable heavy domain (120 residues; UniProt sequence). We restrict mutations to the CDR-H3 segment defined by \citet{mason2021optimization} (the substring WGGDGFYAMD at residues 98-107, zero-indexed inclusive, of the VH); the framework regions, the VL, and the rest of the VH are held fixed.

\paragraph{Beam search} At each step, the 20 children whose realized edit count strictly exceeds their parent's are propagated as the next frontier (so at depth $k$, every frontier sequence has exactly $k$ realized edits). 
Each step also adds the top 200 children to a per-realized-edit-count output pool. 
The final return is the top 200 per realized-edit bucket. We use softmax temperature 1.5 throughout. The realized edit count tracked on each beam node is the Hamming distance to the seed; since we do not allow indels, no alignment is needed. 

\paragraph{Mutation-centric baselines} Gibbs sampling runs for $E$ iterations per chain, sampling one mutable position uniformly at random, masking it, and resampling from the resulting MLM conditional at temperature $\tau = 1$; we run 400 chains per (model, $E$).
Denoising sampling first picks $E$ mutable positions uniformly without replacement, masks all of them simultaneously, and then unmasks one at a time over $E$ iterations; at each iteration we select the masked position whose conditional distribution has the lowest entropy (the ESM-3 default; max-probability is supported but not used here), sample a residue from the conditional at that position, and commit. We run 400 chains per (model, $E$).

\paragraph{Guided variants} Product-of-experts-guided Gibbs adds the supervised oracle's contribution to the MLM masked conditional before sampling. For the binary affinity CNN, the contribution at position $k$ is $\log p_\text{binder}(\bar x; x_k=r)$ for each residue $r$, where $\bar x$ is the current sequence with position $k$ replaced by $r$; we use guidance strength $\gamma = 1$. Beam-search STS guidance combines the MLM pseudo-perplexity (with weight $c_\text{PLL} = 1$) and one or two supervised objectives — the Mason CNN's predicted binder probability and (for the +OASis variants) the OASis percentile — each with weight $c = 2$ in the smooth-Tchebycheff scalarization (matching the recipe used for our in-vitro experiment). 

\paragraph{Affinity oracles}
Both CNNs share the architecture of \citet{mason2021optimization}: \texttt{Conv1D(filters=400, kernel=5, padding=same, ReLU) → Dropout(0.2) → MaxPool1D(pool=2, stride=1) → Flatten → Dense(300, ReLU) → Dense(1, sigmoid)}. Inputs are 10-residue H3 substrings one-hot-encoded into a (10, 20) tensor in alphabetical AA order. We train with binary cross-entropy, Adam (lr 7.5e-5, batch size 16), and early stopping on validation loss with patience 5. We train two: (i) a strong evaluation oracle on the 367k training samples \texttt{her2-aff-large} \citep{chinery2024simple} random split (test ROC AUC 0.995, PR AUC 0.990; used to post-hoc score every generated sequence), and (ii) the Mason CNN guidance oracle on the deduplicated 34k-sample Mason CDR-H3 dataset (test ROC AUC 0.914, PR AUC 0.816 — matching the original reference). 2,342 sequences with conflicting binder/non-binder labels across the original Mason files were dropped before splitting.

\subsection{Additional results for Trastuzumab optimization}
\label{subsec:app-her2-results}

AbLang2 results, presented in Appendix Figure \ref{fig:her2-ablang}, are qualitatively similar with one important twist: the AbLang2 PLL pulls strongly toward non-binder residues, so our unguided beam search performs poorly, like the mutation-centric baselines.
In this scenario, \texttt{MasonCNN} guidance improves Gibbs PoE to \texttt{p(binder)} $>0.5$ at $E=6$, while  \texttt{MasonCNN} guidance improves Beam STS to \texttt{p(binder)} $>0.9$.

\begin{figure}[h!]
    \centering   \includegraphics[width=0.99\linewidth]{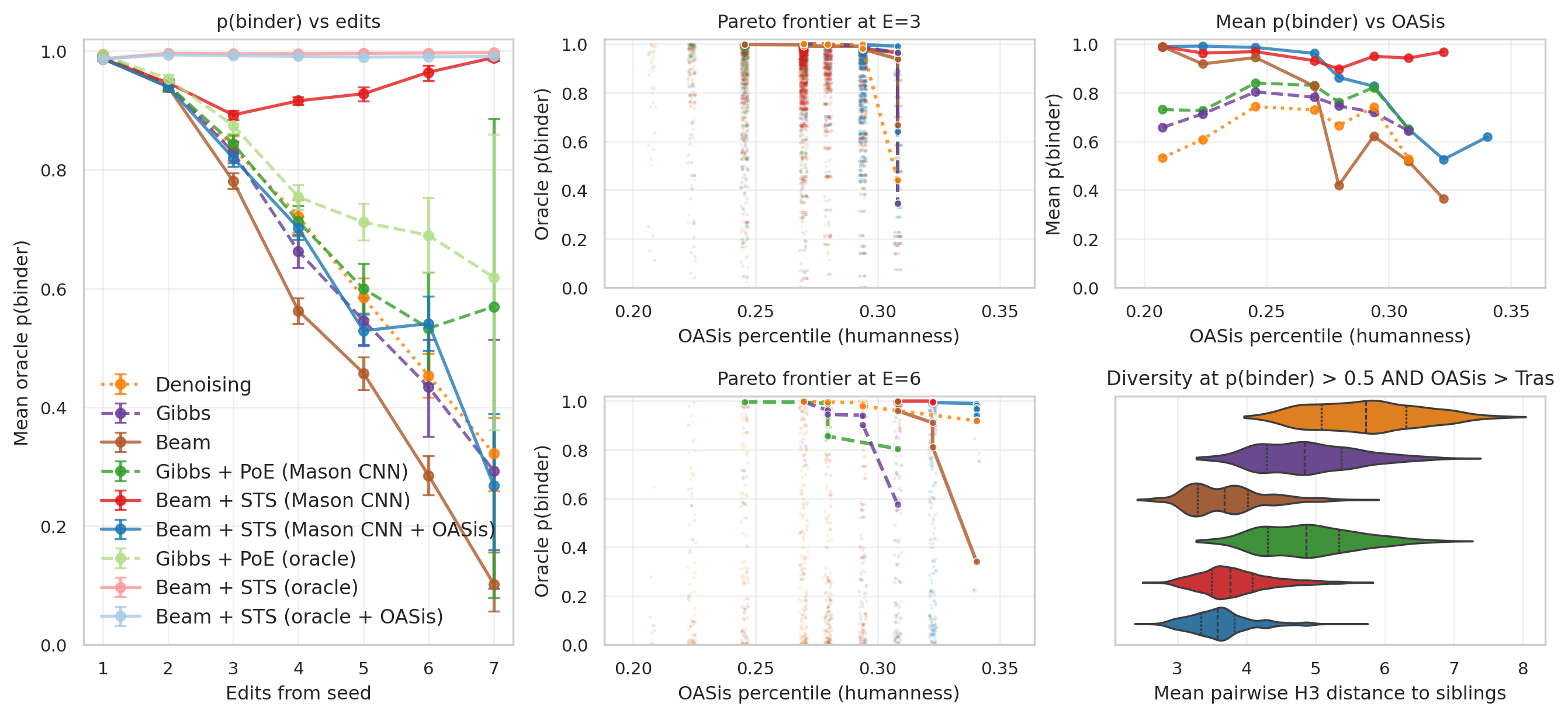}
    \caption{AbLang2 on the Trastuzumab CDR-H3 \textit{in silico} evaluation. Color groups \texttt{MasonCNN}-guided methods with their lighter-shaded strong-oracle counterparts; linestyle indicates methodology (solid: ours; dashed: previous). \textbf{[Left:]} Mean predicted \texttt{p(binder)} vs edits from the seed; error bars depict 95\%-bootstrap CIs. \textbf{[Middle:]} OASis × \texttt{p(binder)} Pareto frontiers. Faded dots show the underlying generated variants. \textbf{[Top Right:]} Mean \texttt{p(binder)} at each OASis percentile bucket, pooling across all edits. \textbf{[Bottom right:]} Pairwise H3-diversity violins restricted to ``high-quality'' variants — those with \texttt{p(binder)} $>$ 0.5 AND OASis $>0.245$, the Trastuzumab seed.}
    \label{fig:her2-ablang}
\end{figure}

\begin{figure}[h!]
    (A) holo (with antigen) \\ 
    \includegraphics[width=0.99\linewidth]{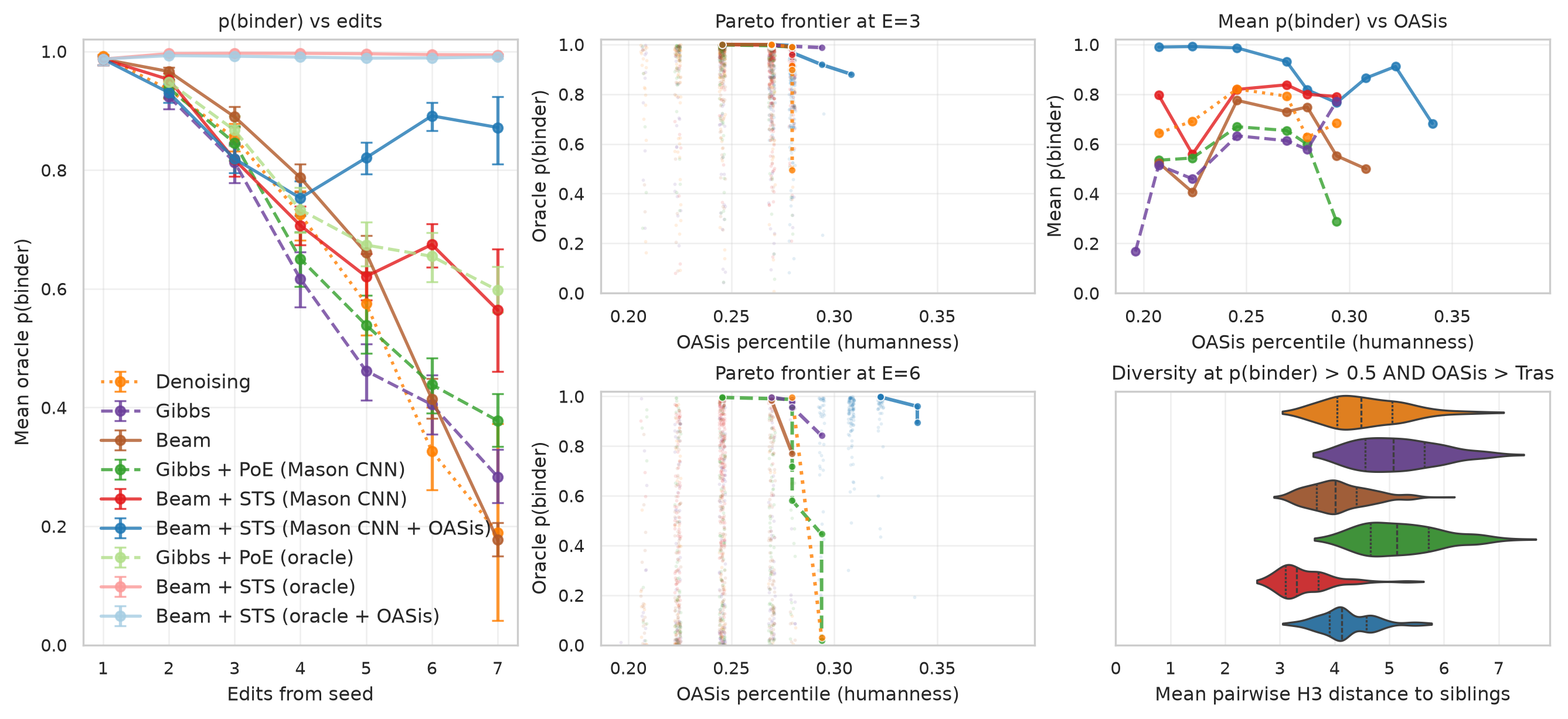}\\
    (B) apo (without antigen) \\
    \includegraphics[width=0.99\linewidth]{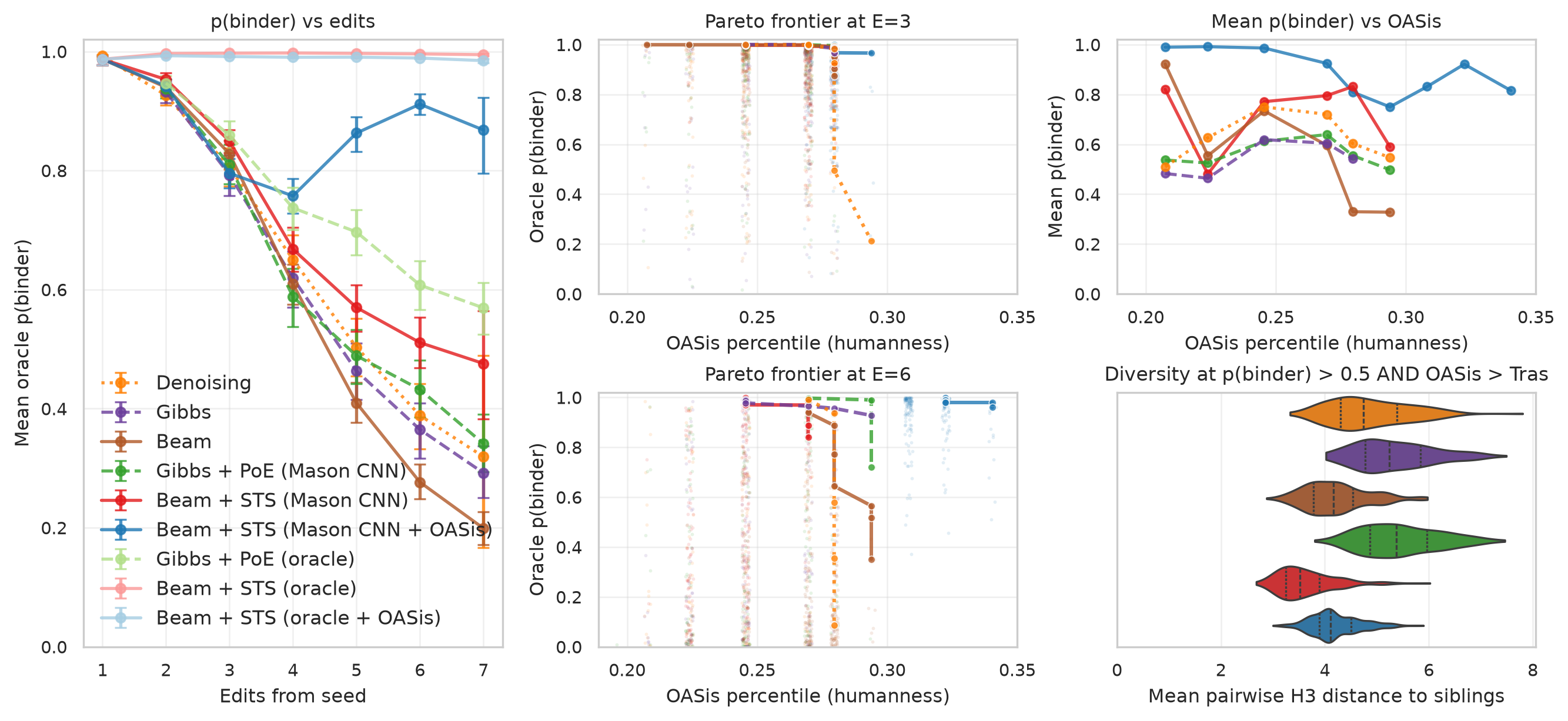}
    \caption{ProteinMPNN on the Trastuzumab CDR-H3 \textit{in silico} evaluation, with (A) and without (B) the bound antigen. Color groups \texttt{MasonCNN}-guided methods with their lighter-shaded strong-oracle counterparts; linestyle indicates methodology (solid: ours; dashed: previous). \textbf{[Left:]} Mean predicted \texttt{p(binder)} vs edits from the seed; error bars depict 95\%-bootstrap CIs. \textbf{[Middle:]} OASis × \texttt{p(binder)} Pareto frontiers. Faded dots show the underlying generated variants. \textbf{[Top Right:]} Mean \texttt{p(binder)} at each OASis percentile bucket, pooling across all edits. \textbf{[Bottom right:]} Pairwise H3-diversity violins restricted to ``high-quality'' variants — those with \texttt{p(binder)} $>$ 0.5 AND OASis $>0.245$, the Trastuzumab seed.}
    \label{fig:her2-proteinmpnn}
\end{figure}

\begin{figure}[h!]
    (A) holo (with antigen) \\ 
    \includegraphics[width=0.99\linewidth]{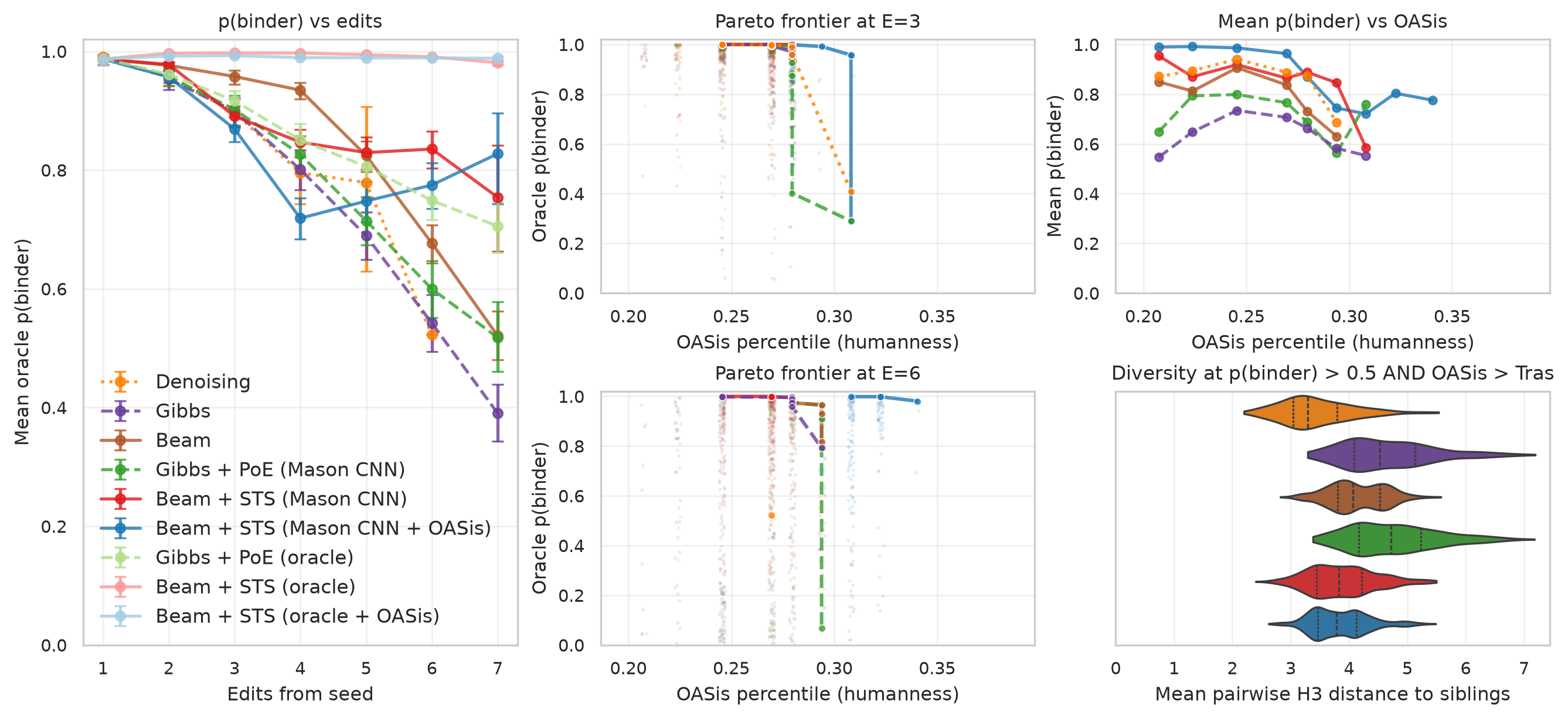}\\
    (B) apo (without antigen) \\
    \includegraphics[width=0.99\linewidth]{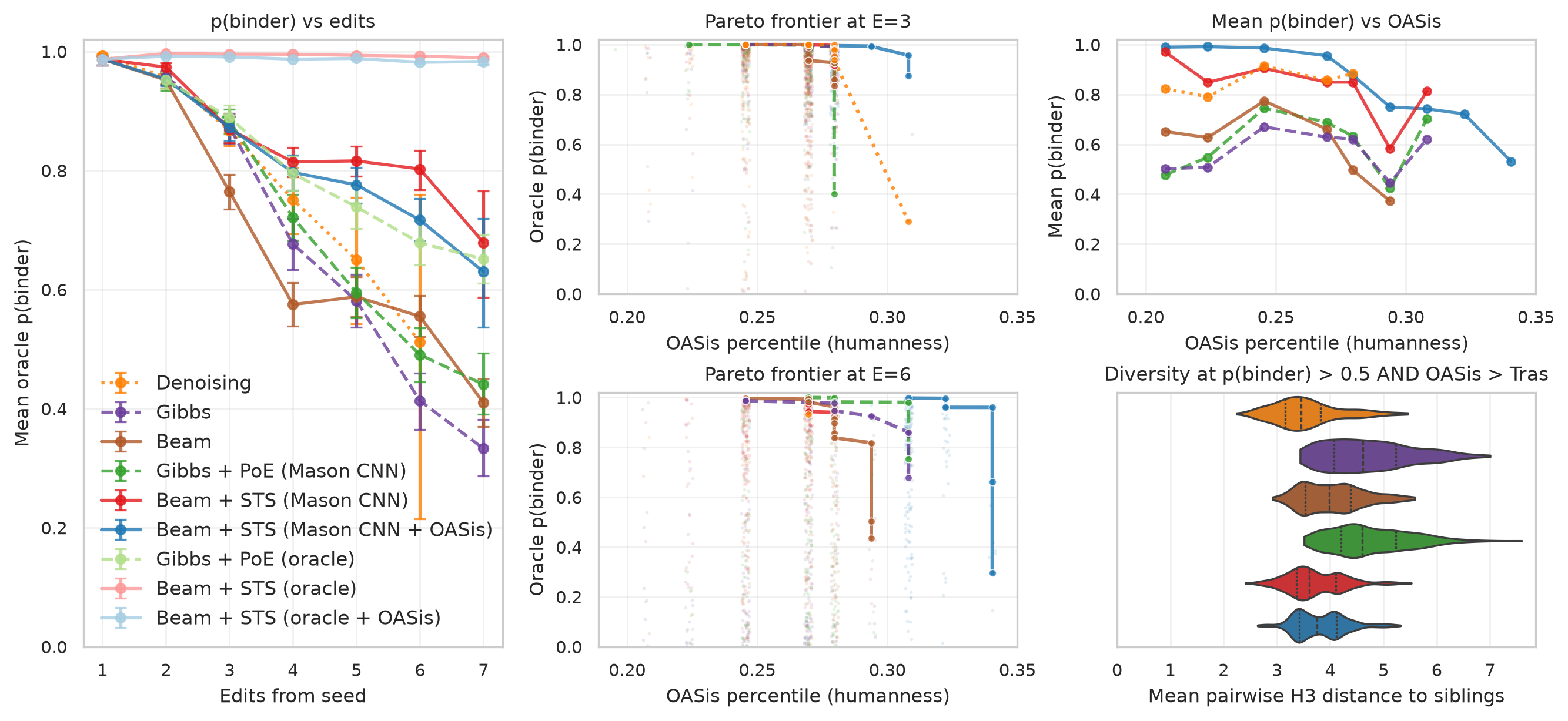}
    \caption{AbMPNN on the Trastuzumab CDR-H3 \textit{in silico} evaluation, with (A) and without (B) the bound antigen. Color groups \texttt{MasonCNN}-guided methods with their lighter-shaded strong-oracle counterparts; linestyle indicates methodology (solid: ours; dashed: previous). \textbf{[Left:]} Mean predicted \texttt{p(binder)} vs edits from the seed; error bars depict 95\%-bootstrap CIs. \textbf{[Middle:]} OASis × \texttt{p(binder)} Pareto frontiers. Faded dots show the underlying generated variants. \textbf{[Top Right:]} Mean \texttt{p(binder)} at each OASis percentile bucket, pooling across all edits. \textbf{[Bottom right:]} Pairwise H3-diversity violins restricted to ``high-quality'' variants — those with \texttt{p(binder)} $>$ 0.5 AND OASis $>0.245$, the Trastuzumab seed.}
    \label{fig:her2-abmpnn}
\end{figure}

The results across all models are summarized in Table \ref{tab:trastuzumab-extended}. 
We see that AbLang2 performs poorly; that ProteinMPNN and AbMPNN have performance strongly dependent on the antigen; that AbMPNN outperforms ProteinMPNN; and that ESM2-650M is at least as good as the inverse-folding MPNNs even with the antigen.

\begin{table}[t]
\centering
\caption{Fraction of designs with oracle $p(\mathrm{binder}) > 0.5$ at $E=6$ edits,
for the extended Trastuzumab models, both with (holo) and without (apo) the
HER2 subdomain IV antigen.}
\label{tab:trastuzumab-extended}
\begin{tabular}{lccccc}
\toprule
Configuration & Beam & Gibbs & Beam:Mason & Gibbs:Mason & Beam:Mason+OASis \\
\midrule
ESM2-650M          & 0.98 & 0.44 & 0.92 & 0.49 & 0.98 \\
AbLang2            & 0.26 & 0.40 & 0.99 & 0.49 & 0.64 \\
ProteinMPNN (holo) & 0.39 & 0.41 & 0.72 & 0.42 & 0.94 \\
ProteinMPNN (apo)  & 0.22 & 0.35 & 0.51 & 0.43 & 0.98 \\
AbMPNN (holo)      & 0.73 & 0.54 & 0.92 & 0.61 & 0.84 \\
AbMPNN (apo)       & 0.57 & 0.40 & 0.88 & 0.47 & 0.78 \\
\bottomrule
\end{tabular}
\end{table}

\subsection{Hyperparameter sweep analysis results}
\label{subsec:app-hyperparameters}

\begin{figure}[h!]
    (A) \\
    \includegraphics[width=0.98\linewidth]{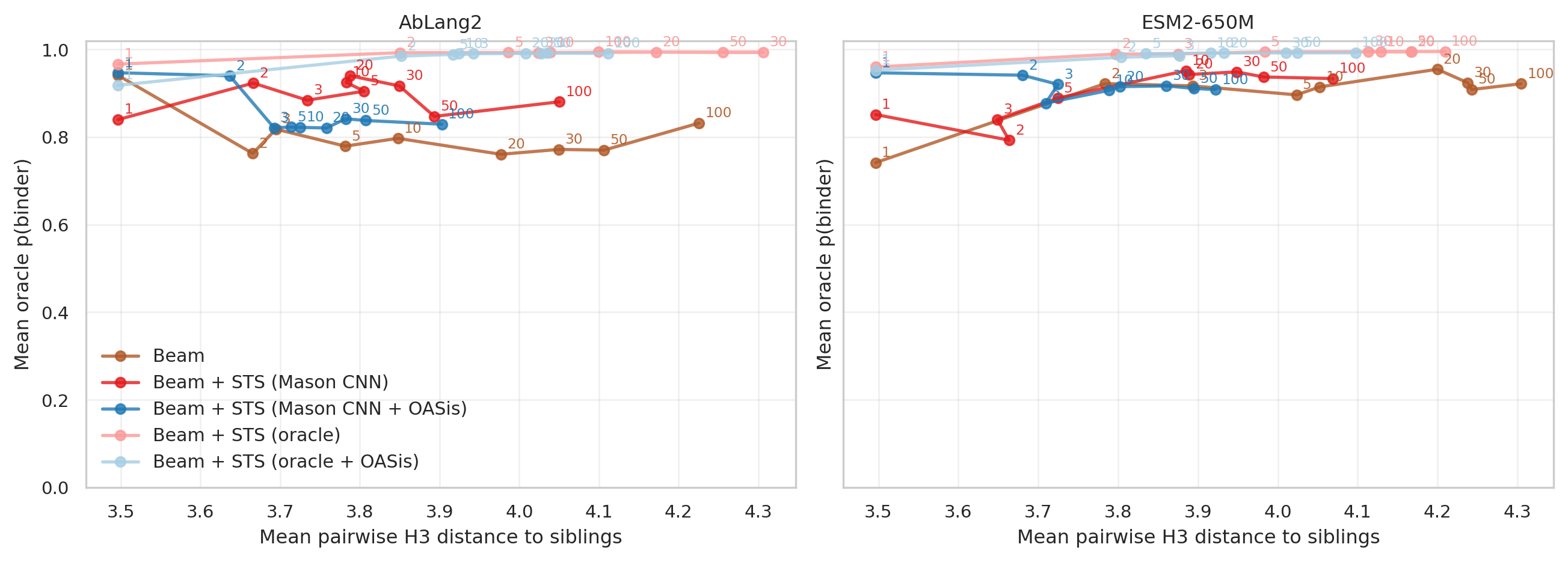} \\
    (B) \\
    \includegraphics[width=0.98\linewidth]{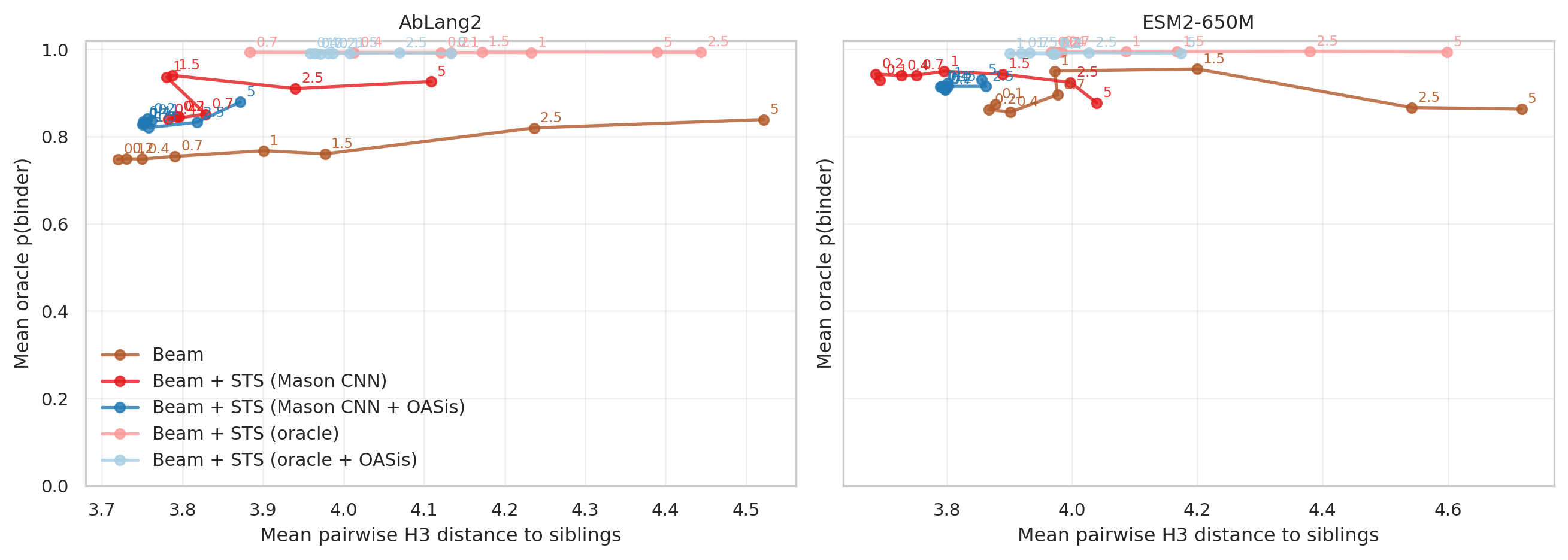} \\
    \caption{Results from sweeping beam size (A) and softmax temperature (B) on Trastuzumab experiment. Text annotations next to each marker denote the hyperparameter used for each experiment.}
    \label{fig:hyperparameter-sweep}
\end{figure}

As shown in Figure \ref{fig:hyperparameter-sweep}, the quality is quite flat as a function of these hyperparameters. Also, one can increase beam size to larger values than used in our experiments, thus increasing diversity, without substantially hurting the quality of generated variants.

\subsection{Runtime analysis results}
\label{subsec:app-runtime}

\begin{figure}[h!]
    \includegraphics[width=0.98\linewidth]{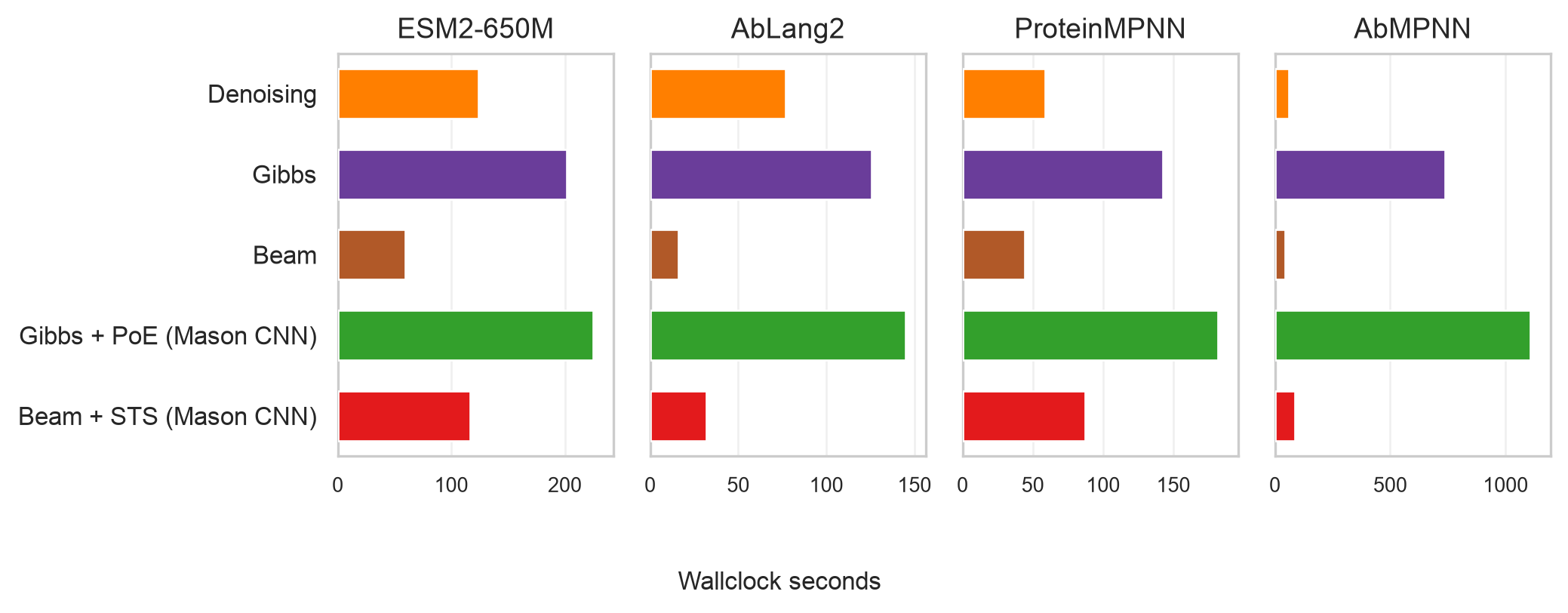}
    \caption{Wallclock time in seconds for Trastuzumab.}
    \label{fig:runtime}
\end{figure}

We measured the wall-clock seconds for the 7-edit Trastuzumab run on an A10G machine, with 1 process on the GPU. Results for models without guidance are shown in Figure \ref{fig:runtime}.
Note that this estimated advantage is smaller than the theoretical advantage, and dramatically underestimates the practical advantage because we designed this particular experiment, for fair quality evaluation, to generate and evaluate the same number of variants, only 200, from Gibbs and Beam. Without any extra cost, for Beam we can actually obtain all $lenMutableRegion \times (numAminoAcids - 1) \times beamSize = 3800$ scored and ranked variants instead (though with the caveats that the lower-ranked variants will be of lower quality and that our scores use the approximate PLL); having all those variants available for downstream ranking and filtering is one of the advantages of our method.

\subsection{Additional details for in silico benchmarking}
\label{subsec:app-insilico}

\paragraph{Experimental setup}

We obtained predicted probabilities of synthesizability (i.e. protein quantification, PQ) from an oracle trained on 6038 samples, split 60\%-20\%-20\% into train-valid-test sets.
We trained a 5-model ensemble of 1d dilated convolutional neural networks \citep{yu2015multi} with mean test AUROC 0.9057.
We obtained melting temperature ($T_m$) predictions by finetuning ESM2-150M \citep{lin2022language} on internal phage display and CFPS data. 
The model was trained to predict first melting temperature (in Celsius) measured by the Uncle (differential scanning fluorimetry) instrument's peak-calling algorithm on the Barycentric Mean (BCM) of intrinsic fluorescence.
For samples from the lineage from which our seeds were drawn, the train-valid-test split was 384-188-219 samples, with test Spearman correlation of 0.7693.
On a larger test-set of 1111 samples from different lineages for the same therapeutics program, the test Spearman was 0.8486.

To compute intra-seed (inter-seed) pairwise diversity, for each child sequence, we align it to all other child sequences of the same (different) seed, and compute the number of mutations.
For each child antibody, we compute the mean number of mutations over all its siblings; we do not aggregate over all children.
Thus, for each method, pairwise diversity is represented by a distribution (each child generated by the method being a sample) rather than a single number.

For computing OASis humanness percentile scores, \citep{prihoda2022biophi} we use the medium ($\ge 50$\% subjects) threshold.
For each child antibody, we computed the germline frequency mutation delta as the sum over all mutated positions (relative to the seed) of the difference between the child residue's germline family frequency and the seed residue's germline family frequency at that position, using IMGT numbering-aware germline gene family frequency tables. 
For computing isoelectric point, we used Biopython v1.85 \citep{cock2009biopython} with the concatenated VH and VL sequences.
For the number of CDR mutations metric, for each child sequence we align \citep{dunbar2016anarci} to the seed, and classify mutations using the Chothia \citep{chothia1987canonical} scheme.

\begin{figure}
    \centering
    \includegraphics[width=0.6\linewidth]{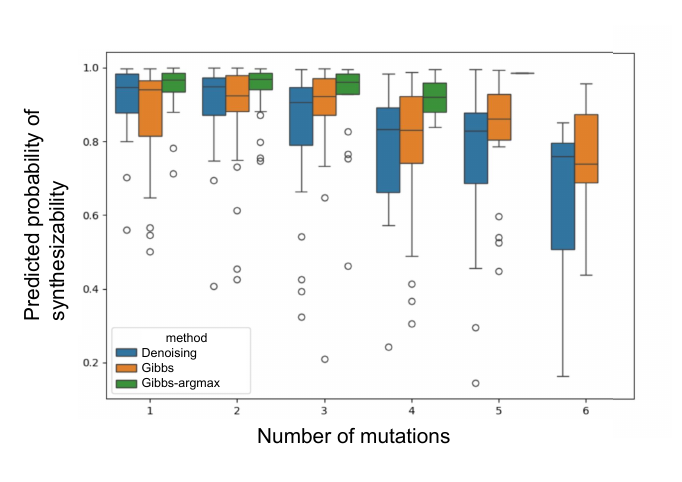}
    \caption{\textit{In silico} predicted probability of synthesizability for AbLang2 generated sequences, as a function of number of edits.}
    \label{fig:ablang-samplers}
\end{figure}

\paragraph{Experimental results}
Results from an initial evaluation of sampling algorithms on AbLang2 are in Figure \ref{fig:ablang-samplers}.

\begin{figure}
    \begin{tabular}{l @{\hspace{1cm}} l}
       (A) & (B) \\
       \includegraphics[width=0.4\linewidth]{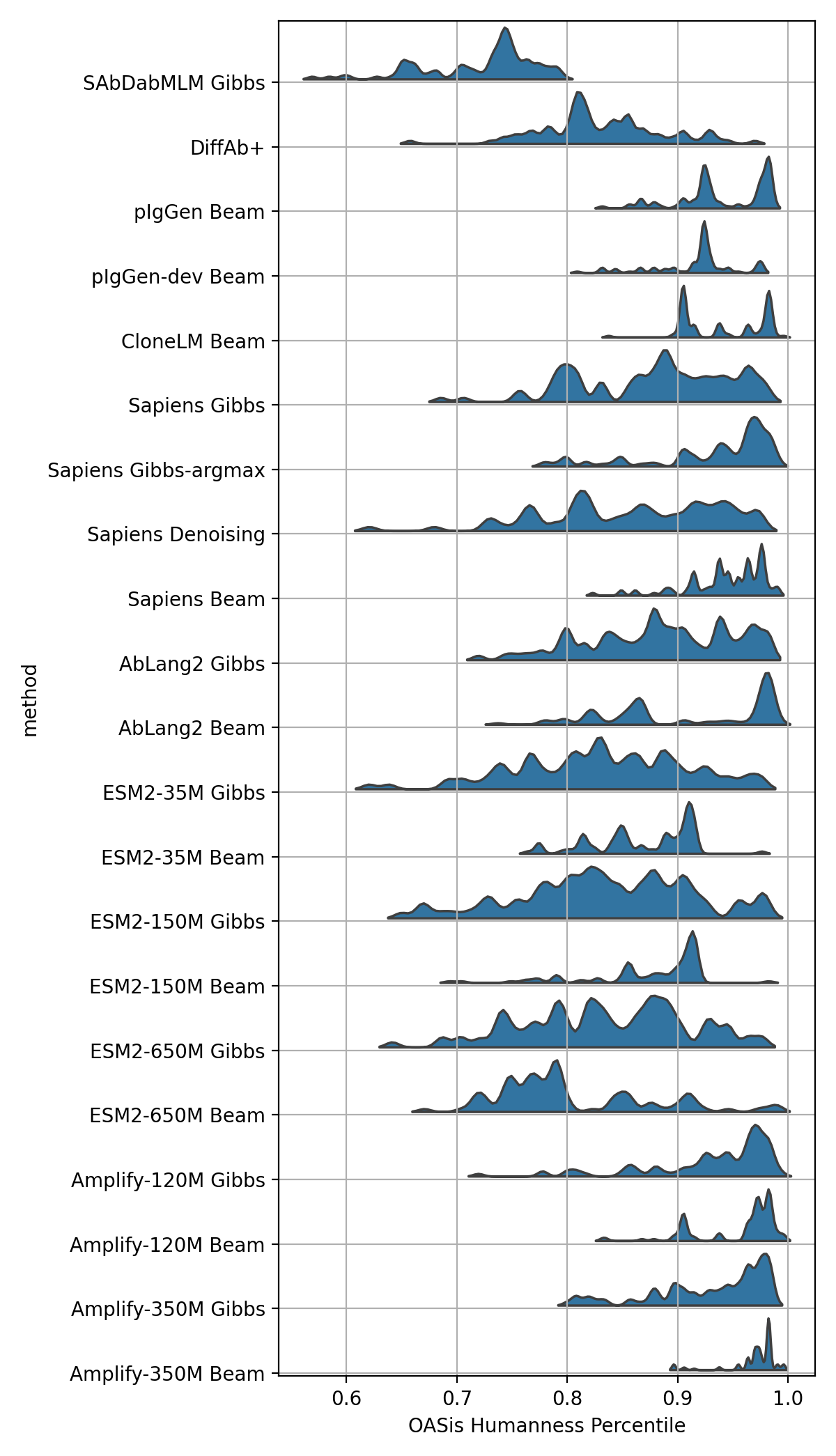}  & \includegraphics[width=0.4\linewidth]{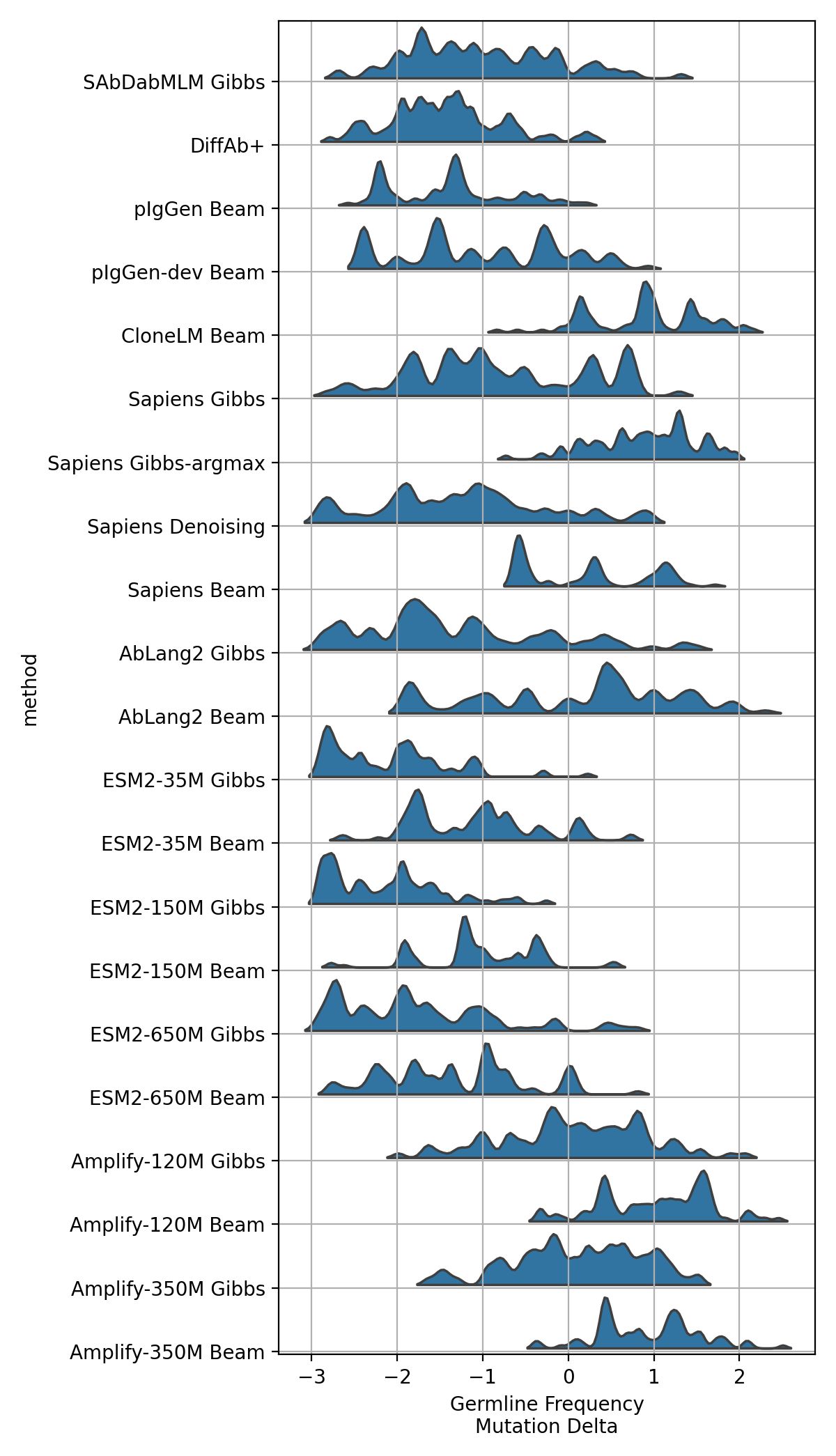} \\
       (C) & (D) \\
       \includegraphics[width=0.4\linewidth]{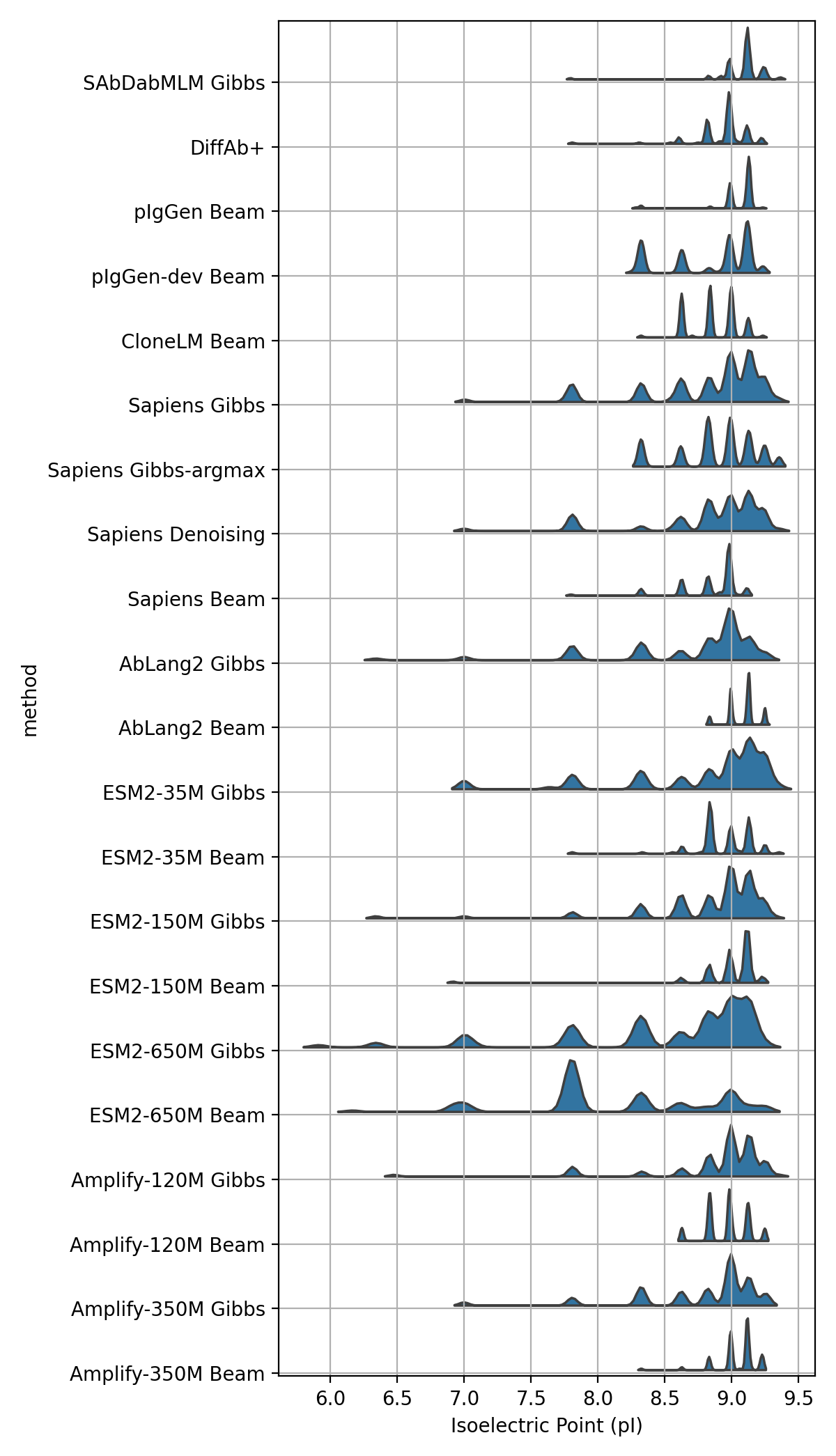}  & \includegraphics[width=0.4\linewidth]{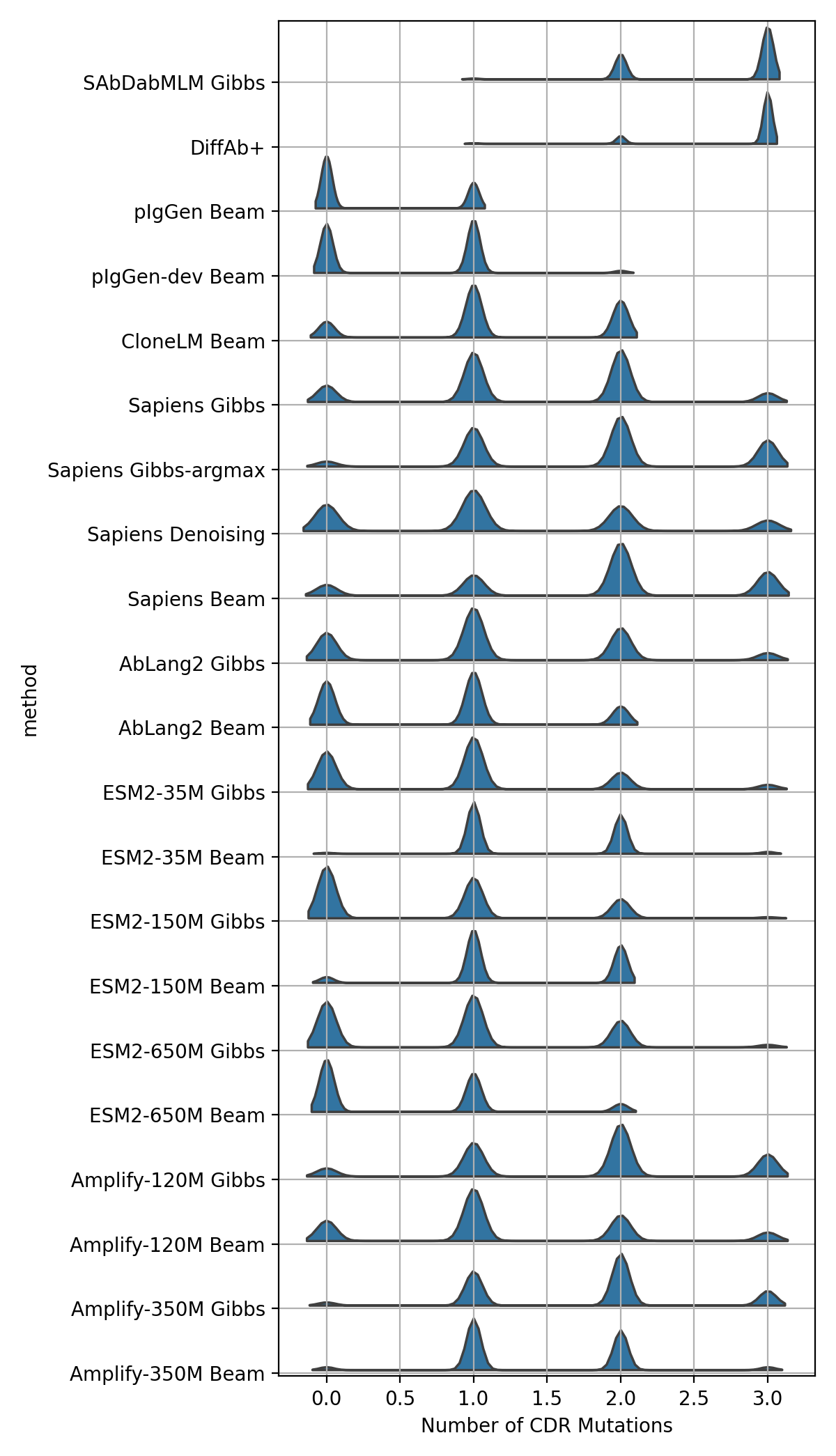} \\       
    \end{tabular}
    \centering    
    \caption{Additional \textit{in silico} evaluation results for generated sequences: (A) OASis humanness percentile, (B) germline mutation bias, (C) isoelectric point, and (D) number of CDR mutations.}
    \label{fig:insilico-more}
\end{figure}

For the full in silico benchmarking results for humanness and germline bias are shown in Fig. \ref{fig:insilico-more}A and Fig. \ref{fig:insilico-more}B, respectively.
The most noteworthy observations are that SAbDabMLM Gibbs produces sequences with low humanness. 
Meanwhile, CloneLM Beam, Sapiens Gibbs-argmax, and Amplify Beam have high germline bias.
For isoelectric point in Fig. \ref{fig:insilico-more}C, we observe that ESM2-650M Beam is the only method that simultaneously performed well in general, while also generating variants with low pI; thus it may be useful in scenarios where pI reduction is needed.
For CDR mutation count in Fig. \ref{fig:insilico-more}D, we see that SAbDabMLM Gibbs is similar to DiffAb+ in being biased to CDR-only mutations, despite not being intentionally designed to do so.
This suggests that this behavior is due to biases in their shared training data.
Conversely, we notice that pIgGen tends to produce mutations in framework regions.

\subsection{Additional details on the in vitro experiment}

\subsubsection{Experimental setup}

When generating variants with CDR-only mutations, we defined CDRs by the union of masks based on the Kabat \citep{kabat1971attempts}, Chothia \citep{chothia1987canonical}, and IMGT \citep{lefranc2003imgt} definitions.

Beam search was run with a beam size of 5, with the exception of the guidance methods which used a beam size of 20.
All methods used a stochastic beam search temperature of 1.5.
For unsupervised methods, the Gumbel-perturbed pseudo-log-likelihoods were used to rank sequences.

For supervised filtering, guidance, and ranking, we split the labeled 729 samples 60\%-20\%-20\% into train-valid-test sets, and trained an ensemble of 5 models.
We aligned sequences to a reference antibody sequence, and provided one-hot encodings of the aligned sequences as input features to the models.
We trained CARP/Bytenet \citep{kalchbrenner2016neural,yang2024convolutions} classification models to predict simultaneous synthesizability and binding success.
The ensemble's models had test mean AUROC 0.9342 and standard deviation 0.0101. 
For filtering, we filtered generated sequences by ($p_{success}>0.6$); for ranking, we sorted generated sequences by $p_{success}$.
For NDS guidance and ranking, during beam search we combined AbLang2 pseudo-perplexity and supervised $p_{success}$, breaking NDS ties with the pseudo-perplexity; at ranking time, sequences were sorted by $p_{success}$.
For STS guidance and ranking, during beam search we combined AbLang2 pseudo-perplexity (with $c_{\textrm{AbLang2}} = 1$) and supervised $p_{success}$ (with $c_{\textrm{supervised}} = 2$); at ranking time, sequences were sorted by $p_{success}$.
Note that at ranking time, the ranker only had access to the top-1000 generated sequences, as sorted by the beam search criteria: for the unguided methods, these are the Gumbel-perturbed pseudo-log-likelihoods; for the guided methods, these are the aforementioned multiobjective scalarization criteria.

Before sending sequences to the wet-lab, designed sequences were filtered to have isoelectric point $\le 9$.
Sequences were also filtered to exclude those introducing any of the following standard sequence liabilities relative to the seed: Asp-Pro cleavage sites, asparagine deamidation motifs (NG, NH, NN, NS, NT), aspartate isomerization motifs (DG, DD, DH, DS, DT), N-linked glycosylation sequons (N-X-S/T, where X $\ne$ P), high-risk oxidation sites (exposed Met or Trp), and unpaired cysteine residues.

We used a seed with -log10 KD of 8.11 (i.e. 7.83 nM), yield of 42.8 ng per microliter, OASis percentile of 16.0\% (14.5\% VH, 17.4\% VL), and overall pI of 6.14 (8.5 VH, 4.87 VL).
For the OASis percentile, it is already OASis identity converted by BioPhi to a percentile against a reference set of therapeutic antibodies drawn from Thera-SAbDab (which includes includes murine and chimeric therapeutics). For pI, we downloaded Thera-SAbDab, then deduplicated and filtered to monospecific and VH:VL-paired, obtaining 914 clinical stage Fv sequences. This has min=4.60, p25=6.85, median=8.30, p75=8.73, and max=9.45.

\subsubsection{In vitro binder classification}

Binding kinetics were measured by bio-layer interferometry (BLI) and fit to a 1:1 binding model. Each sample was subjected to automated quality control to determine whether it produced a reliable binding measurement.       
Before kinetics-specific QC was applied, samples were checked for upstream protein production failures based on protein quantification metrics. A sample was failed if its protein concentration was below 50 ng/µL, or if its A260/A280 purity ratio fell outside the range of 0.5 to 0.85. Samples were classified as non-binders if the maximum binding response signal at the highest analyte concentration was less than 0.05 nm. We manually reviewed select BLI curves, verifying that these automated checks were indeed failing apparent non-binders, while passing apparent binders.

\subsection{Limitations}

The in vitro experimental results come from a single in vitro campaign with a single seed.
These results may not generalize to other antibody campaigns, much less to non-antibody protein campaigns.
Also, in the in vitro experiment, we only evaluated beam search with guidance, not Gibbs with guidance.

\subsection{Computational resources used}

Computations were performed either on a Nov 2023 MacBook Pro (Apple M3 Pro chip, 18GB memory), or on \texttt{g5.4xlarge} or \texttt{g5.12xlarge} AWS EC2 instances.

\newpage

\end{document}